\documentclass[10pt,twocolumn,letterpaper]{article}

\usepackage{iccv}              
\usepackage[accsupp]{axessibility}  

\usepackage{graphicx}
\usepackage{multirow}
\usepackage{colortbl}
\usepackage{amsmath}
\usepackage{array}

\usepackage{amssymb}
\usepackage{amsthm}
\usepackage{booktabs}
\theoremstyle{plain}
\newtheorem{theorem}{Theorem}[section]

\newtheorem{corollary}[theorem]{Corollary}
\theoremstyle{definition}

\theoremstyle{remark}

\definecolor{iccvblue}{rgb}{0.21,0.49,0.74}
\usepackage[pagebackref,breaklinks,colorlinks,allcolors=iccvblue]{hyperref}

\def\paperID{8894} 
\def\confName{ICCV}
\def\confYear{2025}

\title{LoRA-FAIR: Federated LoRA Fine-Tuning with Aggregation and Initialization Refinement}

\author{ Jieming Bian$^{1}$\thanks{The first two authors contributed equally to this work, and their names are listed in random order. } \quad Lei Wang$^{1}$\footnotemark[1] \quad Letian Zhang$^{2}$ \quad Jie Xu$^{1}$ \vspace{0.3em} \\
{\normalsize $^1$ University of Florida }  \quad
{\normalsize $^2$Middle Tennessee State University} \\
{\tt\small jieming.bian@ufl.edu, leiwang1@ufl.edu, letian.zhang@mtsu.edu, jie.xu@ufl.edu }
}

\begin{document}
\maketitle
\begin{abstract}
Foundation models (FMs) achieve strong performance across diverse tasks with task-specific fine-tuning, yet full parameter fine-tuning is often computationally prohibitive for large models. Parameter-efficient fine-tuning (PEFT) methods like Low-Rank Adaptation (LoRA) reduce this cost by introducing low-rank matrices for tuning fewer parameters. While LoRA allows for efficient fine-tuning, it requires significant data for adaptation, making Federated Learning (FL) an appealing solution due to its privacy-preserving collaborative framework. However, combining LoRA with FL introduces two key challenges: the \textbf{Server-Side Aggregation Bias}, where server-side averaging of LoRA matrices diverges from the ideal global update, and the \textbf{Client-Side Initialization Lag}, emphasizing the need for consistent initialization across rounds. Existing approaches address these challenges individually, limiting their effectiveness. We propose LoRA-FAIR, a novel method that tackles both issues by introducing a correction term on the server, enhancing aggregation efficiency and accuracy. LoRA-FAIR maintains computational and communication efficiency, yielding superior performance over state-of-the-art methods. Experimental results on ViT and MLP-Mixer models across large-scale datasets demonstrate that LoRA-FAIR consistently achieves performance improvements in FL settings. 
\end{abstract}
    
\section{Introduction}
\label{sec:intro}

\begin{figure}[t]
	\centering	
    \includegraphics[width=0.86\linewidth]{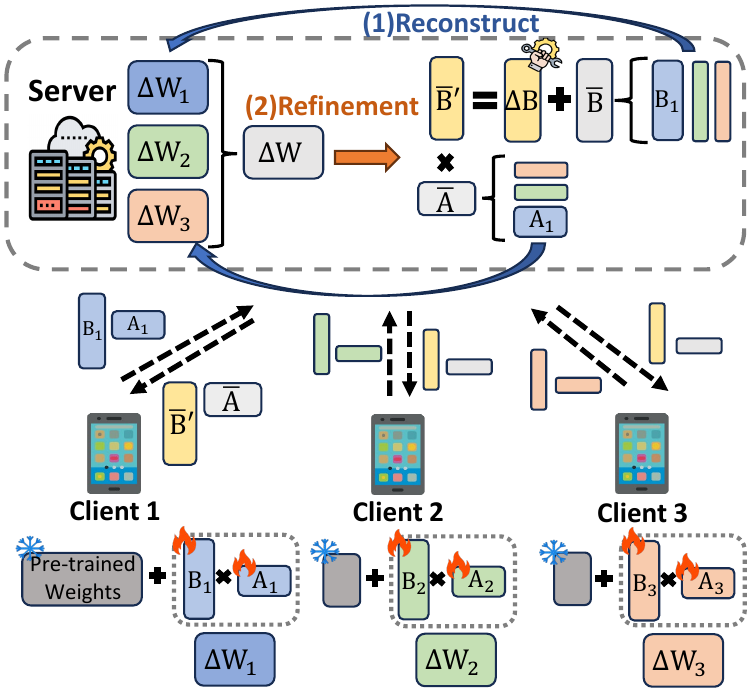}
	\caption{\textbf{Illustration of LoRA-FAIR.} Instead of directly averaging the local LoRA modules $\mathbf{A}_k$ and $\mathbf{B}_k$ collected from each client $k$ on the server side and sending the averaged LoRA modules $\mathbf{\Bar{A}}$ and $\mathbf{\Bar{B}}$ back to clients, LoRA-FAIR reconstructs the ideal global update $\mathbf{\Delta W}$ using \cref{eq:ideal_update}, finds the residual LoRA module $\mathbf{\Delta B}$ using \cref{eq:optimization_problem}, and replaces $\mathbf{\Bar{B}}$ with the corrected LoRA modules $\mathbf{\Bar{B}'} = \mathbf{\Bar{B}} + \mathbf{\Delta B}$. See details in \cref{sec:method}.}
    \vspace{-20pt}
    \label{fig: lorafair}
\end{figure}

Emerging foundation models (FMs) \cite{achiam2023gpt, bommasani2021opportunities, touvron2023llama, zeng2022glm, zhou2023comprehensive} have demonstrated remarkable capabilities by providing robust and versatile architectures that can be adapted to a wide array of tasks through fine-tuning with task-specific data. These models excel across diverse applications, including image generation from prompts, language translation, mathematical problem-solving, and natural language conversation, among others \cite{zhou2023comprehensive}. However, the standard method of fine-tuning all model parameters, known as full parameter fine-tuning, entails prohibitively high computational costs, particularly for large-scale models. To alleviate this problem, parameter-efficient fine-tuning (PEFT) methods \cite{han2024parameter} have been proposed. One of the most important PEFT approaches is low-rank adaptation (LoRA) \cite{hu2021lora}, which significantly reduces the number of trainable parameters by introducing low-rank matrices into the model. 

LoRA introduces a parallel branch of trainable low-rank matrices, \(\mathbf{A}\) and \(\mathbf{B}\), to compute the model update \(\mathbf{\Delta W}\), where the ranks of \(\mathbf{A}\) and \(\mathbf{B}\) are significantly smaller than the parameters of the pre-trained model, \(\mathbf{W}\). In LoRA fine-tuning, only \(\mathbf{A}\) and \(\mathbf{B}\) are updated, while \(\mathbf{W}\) remains frozen. This approach greatly reduces the computational resources required, allowing for efficient fine-tuning with performance comparable to that of full parameter fine-tuning. Despite these advantages, LoRA still requires substantial data to adapt effectively to specific downstream tasks. However, data from a single device may not be sufficient for this purpose, and fine-tuning often involves multiple devices that collectively hold the necessary data. This multi-device setup can raise privacy concerns, as fine-tuning with data from multiple parties may expose sensitive information. Federated Learning (FL) \cite{mcmahan2017communication} offers a feasible solution to this issue. By enabling collaborative learning without requiring data sharing, FL allows participants to fine-tune models while addressing privacy concerns effectively.


Compared to studies on LoRA fine-tuning in centralized settings, fine-tuning LoRA within a FL environment remains relatively unexplored and presents unique challenges. In this paper, we investigate traditional FL in conjunction with parameter-efficient fine-tuning methods, specifically focusing on LoRA. We argue that fine-tuning LoRA modules presents two key challenges. First, which we refer to as the \textbf{Challenge 1: Server-Side Aggregation Bias}, arises because averaging the LoRA components (\( \mathbf{A} \) and \( \mathbf{B} \)) independently at the server does not capture the ideal global update, potentially introducing noise into the aggregated model. Second, \textbf{Challenge 2: Client-Side Initialization Lag} highlights the importance of properly allocating global updates to each client's pre-trained model and LoRA modules at the start of the next local training phase to ensure a consistent initialization and mitigate initialization lag. Existing FL methods for fine-tuning fail to consider these two key points simultaneously. While some methods, such as FLoRA \cite{wang2024flora}, attempt to address Challenge 1 by altering the aggregation process, they fail to address Challenge 2, which limits the performance to a level comparable to that of directly combining FedAvg and LoRA (i.e., FedIT \cite{zhang2024towards}).

Taking both Challenge 1 and Challenge 2 into consideration simultaneously is essential for maximizing the performance of LoRA fine-tuning in a federated learning setting. In this work, we propose a simple yet effective method, LoRA-FAIR (short for LoRA with \underline{F}ederated \underline{A}ggregation and \underline{I}nitialization \underline{R}efinement), designed to tackle both challenges concurrently. Specifically, we propose that, on the server side, the original averaged LoRA modules (e.g., $\mathbf{\bar{A}}$ and $\mathbf{\bar{B}}$) be kept fixed while introducing a correction term $\mathbf{\Delta B}$ to $\mathbf{\bar{B}}$. This way, the product of the fine-tuned \( \mathbf{\bar{B}} + \mathbf{\Delta B}  \) and \( \mathbf{\bar{A}} \) will closely approximate the ideal server update. To further enhance stability, we introduce a normalization term to ensure that the fine-tuned LoRA module remains close to its original averaged value, thereby preserving the average information collected from each client. Through this simple yet effective design, LoRA-FAIR provides an approach that approximates an ideal solution to both challenges by preserving the shared average information in the initial model while striving for accurate aggregation on the server side. Consequently, LoRA-FAIR maximizes the efficacy of LoRA fine-tuning within an FL framework, balancing performance improvements with computational efficiency. Our key contributions are summarized as follows:

\begin{itemize} 
\item We investigate the problem of fine-tuning with LoRA in federated learning setting. Through an initial set of motivation experiments, we identify two key challenges that currently limit the application of LoRA in FL. 
\item In response to these challenges, we introduce a novel method named LoRA-FAIR. LoRA-FAIR is the first in the federated fine-tuning domain to simultaneously consider both the two challenges while maintaining computational and communication efficiency. 
\item We conduct experiments using two pre-trained foundation models, ViT \cite{dosovitskiy2020image} and MLP-Mixer \cite{tolstikhin2021mlp}, across various large-scale datasets. Our proposed LoRA-FAIR consistently outperforms state-of-the-art methods. 
\end{itemize}

\section{Preliminaries}

\subsection{PEFT with LoRA}
LoRA (Low-Rank Adaptation) is a PEFT (parameter-efficient fine-tuning) approach that significantly reduces the number of trainable parameters in large-scale models by introducing low-rank matrices into the model. Consider a pre-trained model with parameters $\mathbf{W}_0 \in \mathbb{R}^{d \times l}$, where $\mathbf{W}_0$ represents the fixed parameters of the model, and $\mathbf{\Delta W} \in \mathbb{R}^{d \times l}$ denotes the trainable update matrix applied during fine-tuning. Rather than updating all elements in $\mathbf{\Delta W}$, LoRA decomposes $\mathbf{\Delta W}$ into two low-rank matrices $\mathbf{A} \in \mathbb{R}^{r \times l}$ and $\mathbf{B} \in \mathbb{R}^{d \times r}$, where $r \ll \min(d, l)$. Thus, the model update is expressed as $\mathbf{\Delta W} = \mathbf{B} \mathbf{A}$, allowing the fine-tuning process to focus on the much smaller low-rank matrices $\mathbf{A}$ and $\mathbf{B}$ instead of the full matrix $\mathbf{\Delta W}$. Consequently, the total number of parameters that need to be trained is reduced from $d \times l$ to $r \times (d + l)$, where $r$ is significantly smaller than both $d$ and $l$. The updated model parameters after fine-tuning are given by:

\begin{equation}\label{lora} \mathbf{W} = \mathbf{W}_0 + \mathbf{\Delta W} = \mathbf{W}_0 + \mathbf{B} \mathbf{A}. \end{equation}

In practice, $\mathbf{A}$ is typically initialized with random Gaussian values, while $\mathbf{B}$ is initialized to zero to ensure a stable start to the fine-tuning process. This low-rank adaptation enables LoRA to achieve performance comparable to full fine-tuning while significantly reducing the computational and memory overhead.

\subsection{Federated Learning}

In a standard federated learning setup, multiple clients collaboratively train a shared global model without sharing their local data, thereby preserving privacy. Each client trains on its local data and then transmits its local model updates back to the server, which aggregates these updates to refine the global model. 

Consider an FL setup with \( K \) clients, starting with an initial model \( \mathbf{W}_0 \). The server collects the local updates from the clients and calculates the global update as follows:

\begin{align}
\label{fl_update}
    \mathbf{\Delta W} = \sum_{k=1}^K p_k \mathbf{\Delta W}_k,
\end{align}
where $\mathcal{D}_k$ is the client $k$'s local dataset, the weights \( p_k = \frac{|\mathcal{D}_k|}{\sum_{k} |\mathcal{D}_k|} \) are proportional to the size of each client’s local dataset, and \( \mathbf{\Delta W}_k \) denotes the local update from client \( k \). To start the next round of local training, the server uses the global update \( \mathbf{\Delta W} \) to generate an updated global model, which is then distributed to each client as the initial model for the subsequent round. The next round of training for each client can be represented as follows, assuming clients train for \( E \) epochs during local training:

\begin{align}
\label{fl_initial}
    &\mathbf{W}_{k,0} = \mathbf{W}_0 + \mathbf{\Delta W}; \nonumber\\
    &\mathbf{W}_{k,e+1} = \mathbf{W}_{k,e} - \eta g_{k,e}, \quad e = 0, \dots, E-1; \nonumber\\
    &\mathbf{\Delta W}_k = -\sum_{e=0}^{E-1} \eta g_{k,e},
\end{align} 
where \( \eta \) is the local learning rate, and \( g_{k,e} \) represents the stochastic gradient for client \( k \) at epoch \( e \).

\section{Challenges when Combining LoRA with Federated Learning}
\label{sec:challenge}
Fine-tuning foundation models in federated learning using full-parameter updates aligns with traditional FL methods. However, incorporating LoRA introduces unique challenges that diverge from those in centralized settings.

\subsection{Challenge 1: Server-Side Aggregation Bias}
\label{sec:challenge1}
\begin{figure}[t]
	\centering
   	 	\includegraphics[width=0.9\linewidth]{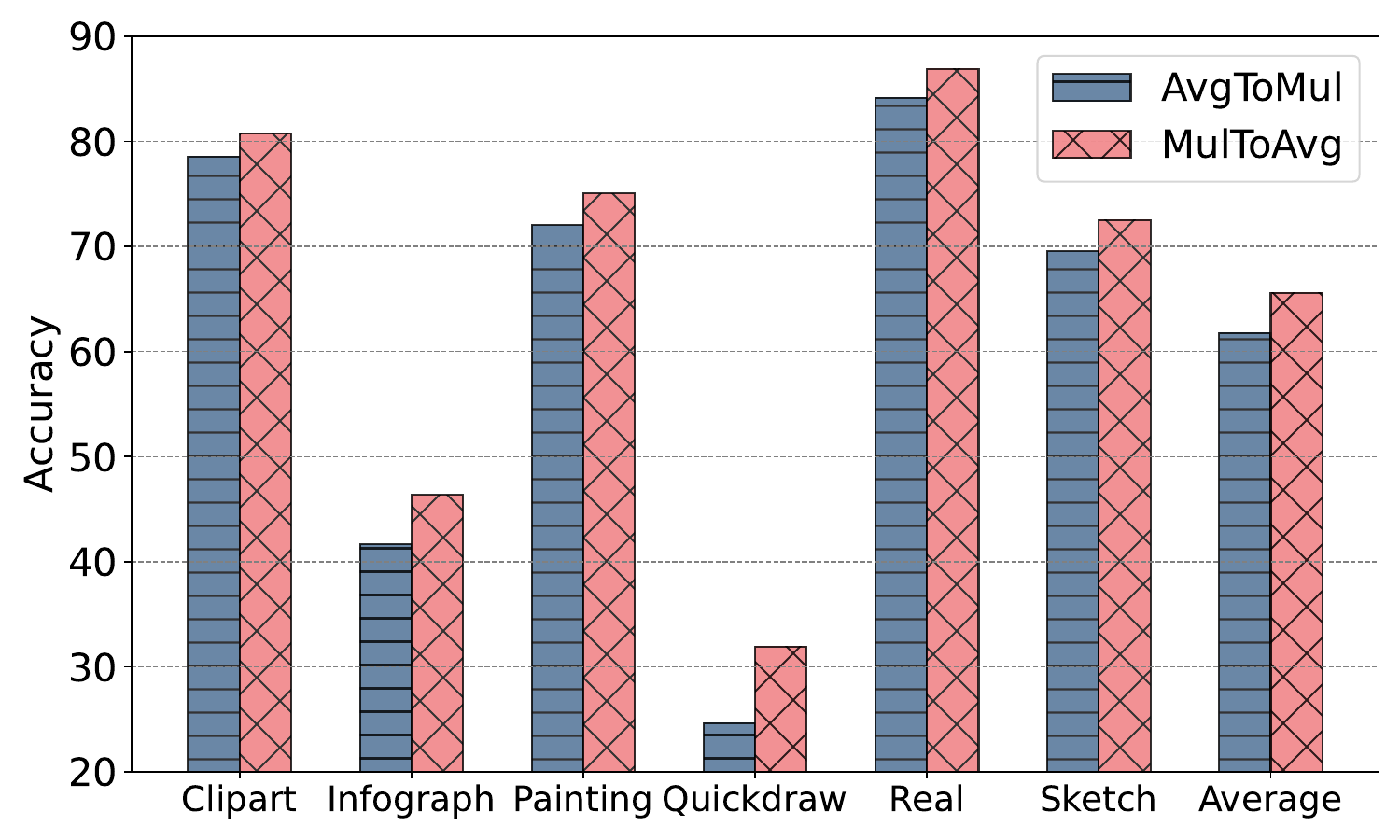}
   	\caption{\textbf{Comparison of two aggregation strategies: AvgToMul and MulToAvg.} \textbf{AvgToMul} averages the LoRA matrices $\mathbf{A}_k$ and $\mathbf{B}_k$ from clients, then multiplies the averages to obtain the approximate global update $\mathbf{\Delta W}'$ using \cref{eq:approx_update}. \textbf{MulToAvg} first multiplies each client’s matrices (yielding $\mathbf{B}_k \mathbf{A}_k$) and then averages these products for the true global update $\mathbf{\Delta W}$ using \cref{eq:ideal_update}. While \textbf{AvgToMul} is communication-efficient, \textbf{MulToAvg} better captures the intended global model update. See details in \cref{sec:challenge1}.}
   	\label{fig:c1}
	 \vspace{-10 pt}
\end{figure}

To discuss this challenge, we first introduce a basic method that combines LoRA directly with FL, known as FedIT \cite{zhang2024towards}. In FedIT, each of the $K$ clients starts with a fixed pre-trained foundation model $\mathbf{W}_0$ and trains the local LoRA modules represented as low-rank matrices \( \mathbf{A}_k \) and \( \mathbf{B}_k \) on its private dataset $\mathcal{D}_k$. The server then aggregates these local matrices uploaded by clients into global LoRA modules, $\mathbf{\bar{A}}$ and $\mathbf{\bar{B}}$, through a weighted average based on data size:
\begin{equation}
\label{global_modules}
\mathbf{\bar{A}} = \sum_{k=1}^K p_k \mathbf{A}_k, \quad \mathbf{\bar{B}} = \sum_{k=1}^K p_k \mathbf{B}_k,
\end{equation}
where \( p_k = \frac{|\mathcal{D}_k|}{\sum_{k=1}^K |\mathcal{D}_k|} \) reflects each client's data proportion. Using these averaged matrices, the server distributes them back to the clients for subsequent training rounds. In FedIT, the actual global update received by each client is:
\begin{equation}
\label{eq:approx_update}
\mathbf{\Delta W}' = \mathbf{\bar{B}} \mathbf{\bar{A}} = \left( \sum_{k=1}^K p_k \mathbf{B}_k \right) \left( \sum_{k=1}^K p_k \mathbf{A}_k \right).
\end{equation}
However, this aggregated update deviates from the ideal global model update in the typical FL setting, which should be the weighted sum of all local model updates:
\begin{equation}
\label{eq:ideal_update}
\mathbf{\Delta W} = \sum_{k=1}^K p_k \mathbf{\Delta W}_k = \sum_{k=1}^K p_k \mathbf{B}_k \mathbf{A}_k \neq \mathbf{\Delta W}'.
\end{equation}
This discrepancy, termed \textbf{Server-Side Aggregation Bias}, occurs because the approximate global update $\mathbf{\Delta W}'$ fails to accurately capture the ideal global update $\mathbf{\Delta W}$. To demonstrate this, we compare the two aggregation methods under a single global round with 50 local epochs independent of the client-side initialization on the DomainNet dataset. As shown in \cref{fig:c1}, \textbf{AvgToMul} and \textbf{MulToAvg} denotes the aggregated update using $\mathbf{\Delta W}'$ and $\mathbf{\Delta W}$ respectively. Although \textbf{AvgToMul} reduces communication costs by only transmitting the LoRA modules, it does so at the expense of alignment with the intended global model update. This challenge highlights the need for more refined aggregation methods when integrating LoRA into FL frameworks.

\subsection{Challenge 2: Client-Side Initialization Lag}
\label{sec:challenge2}


\begin{figure}[t]
	\centering
	\includegraphics[width=0.9\linewidth]{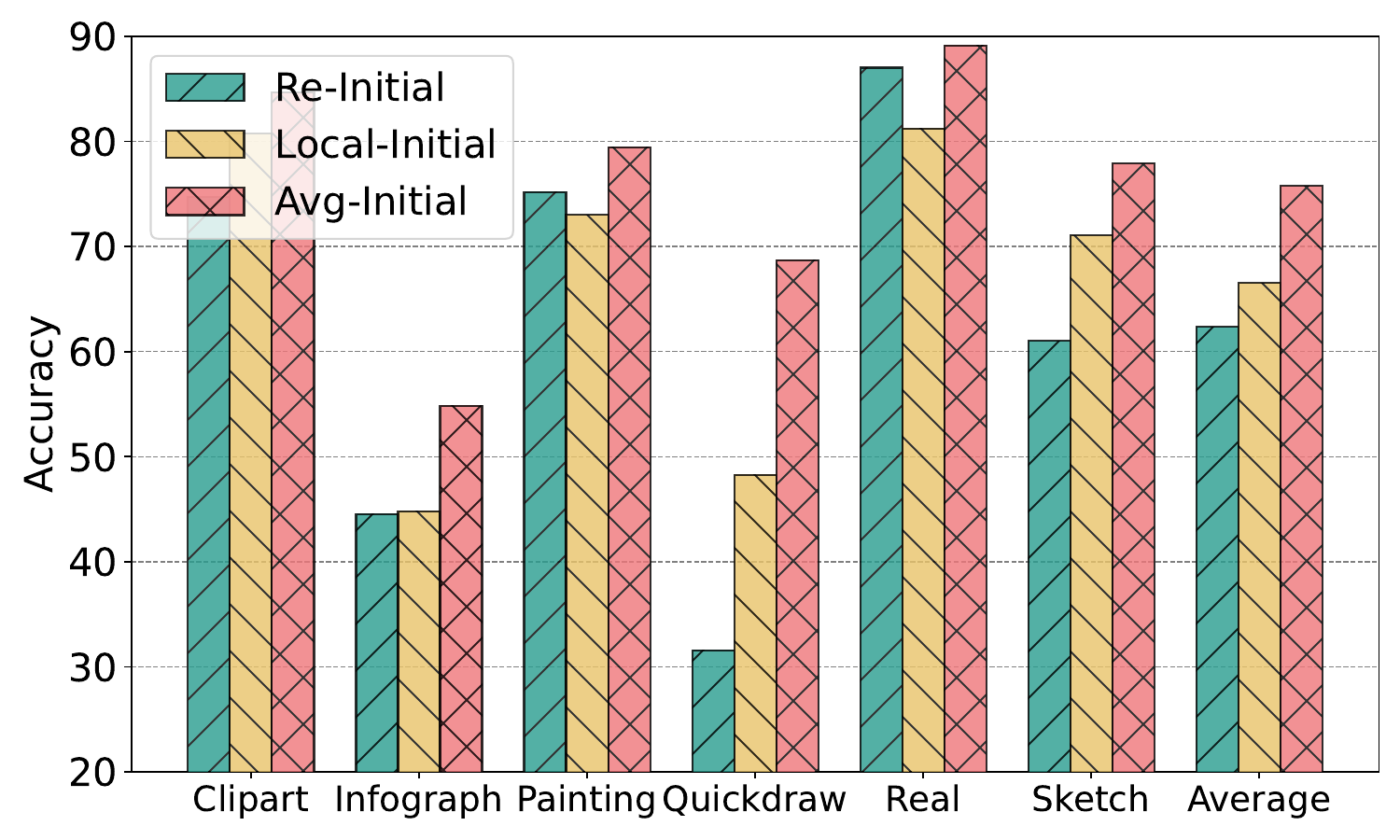} 
   	\caption{\textbf{Comparison of three initialization strategies: Avg-Initial, Re-Initial, Local-Initial.} The \textbf{Avg-Initial} method is the most effective as it balances continuity and unification across clients, mitigating client initialization lag and promoting better performance. For more details, refer to \cref{sec:challenge2}.}
   	\label{fig:c2}
	\vspace{-15pt} 
\end{figure}

To mitigate server-side aggregation bias, FFA-LoRA~\cite{sun2024improving} was proposed, which freezes the non-zero-initialized low-rank matrix $\mathbf{A}$ while updating only the zero-initialized matrix $\mathbf{B}$. However, this approach slows fine-tuning and limits performance due to the reduced number of trainable parameters. A more recent method, FLoRA~\cite{wang2024flora}, stacks local LoRA modules from all clients and transmits the aggregated modules back to each client to reconstruct global updates. These updates are then added directly to each local pre-trained model, while the local LoRA modules are reinitialized for the next training round. FLoRA effectively addresses \textbf{Challenge 1} by stacking local LoRA modules, ensuring that each client receives an ideal $\mathbf{\Delta W}$ update to add to the pre-trained model. However, this method incurs high communication costs proportional to the number of clients and raises privacy concerns, as it requires distributing all clients' LoRA modules to each client rather than only the averaged modules, as in FedIT.

Furthermore, in FLoRA, the client's local LoRA modules are reinitialized (randomizing $\mathbf{A}$ with a Gaussian distribution and setting $\mathbf{B}$ to zero). This reinitialization strategy can lead to \textbf{Client-Side Initialization Lag}. Given an input $x$ and output $y$ at a layer, a forward pass with LoRA modules is represented as: $   y = x(\mathbf{W}_0 + \mathbf{B} \mathbf{A})$.
Accordingly, the gradients of $\mathbf{A}$ and $\mathbf{B}$ are:
\begin{equation}
    \frac{\partial L}{\partial \mathbf{A}} = x^\top \frac{\partial L}{\partial y} \mathbf{B}^\top, \quad 
    \frac{\partial L}{\partial \mathbf{B}} = \mathbf{A}^\top x^\top \frac{\partial L}{\partial y}.
\end{equation}
When $\mathbf{A}$ is initialized with Gaussian noise and $\mathbf{B}$ is set to zero, the initial gradients are small and uninformative (i.e., $\frac{\partial L}{\partial \mathbf{A}} \rightarrow 0, \frac{\partial L}{\partial \mathbf{B}} \rightarrow \textit{random direction}$), leading to a slow learning start. LoRA then spends significant time near its initialization before meaningful updates occur~\cite{meng2025pissa}. In FL setting, where clients often have non-IID data, a prolonged local training phase can exacerbate global convergence issues ~\cite{zhao2018federated}. Under a limited number of local training epochs, reinitialization may prevent the model from effectively capturing optimal local updates. Consequently, the locally learned information sent to the server may be \textbf{\textit{suboptimal}}, which in turn degrades the global model’s performance—despite FLoRA’s ability to mitigate server-side aggregation bias.

\begin{table}[h]
\centering
\resizebox{\columnwidth}{!}{%
\begin{tabular}{c|ccc|c}
\toprule
    Strategies & $\mathbf{W_0} \gets$ & $\mathbf{A_k} \gets$ & $\mathbf{B_k} \gets$ & Overall Initial Model \\
\midrule
Avg-Initial & $\mathbf{W_0}$ & $\bar{\mathbf{A}}$ & $\bar{\mathbf{B}}$ & $\mathbf{W_0 + \Delta W'}$ \\
Re-Initial & $\mathbf{W_0} + \Delta \mathbf{W}'$ & \textit{Random Gaussian} & $\mathbf{0}$ & $\mathbf{W_0 + \Delta W'}$ \\
Local-Initial & $\mathbf{W_0} + \Delta \mathbf{W}' - \mathbf{B_s}\mathbf{A_s}$ & $\mathbf{A_s}$ & $\mathbf{B_s}$ & $\mathbf{W_0 + \Delta W'}$ \\
\bottomrule
\end{tabular}%
}
\caption{\textbf{Different Client Initialization Strategies.} Note that $\mathbf{\Delta W}' = \bar{\mathbf{B}}\bar{\mathbf{A}}$ is reconstructed locally. $\mathbf{B_s}$ and $\mathbf{A_s}$ represent a randomly selected client's last-round local LoRA modules. See details in \cref{sec:challenge2}.}
\label{tab: initialization}
\vspace{-15pt}
\end{table}

This raises the question: \textit{what is the optimal way for a client to allocate the received global update to mitigate Client-Side Initialization Lag?} To evaluate the impact of different client-side initialization methods on model performance, we consider three strategies in an FL setup with six clients, each assigned a unique domain from the DomainNet dataset. To isolate the effect of initialization strategies from server-side aggregation, all clients receive the bias $\mathbf{\Delta W}'$ from the server in different formats based on the applied strategies. The three strategies are described in \cref{tab: initialization}. \textbf{Local-Initial} requires an additional randomly selected client's last-round local LoRA modules $\mathbf{A_s}, \mathbf{B_s}$ as the LoRA initialization point for all clients. Although all approaches result in the same overall initial model (i.e. $\mathbf{W_0} + \mathbf{B_k}\mathbf{A_k} \gets \mathbf{W_0} + \Delta\mathbf{W'}$) at the start of the current training round, as shown in \cref{fig:c2}, the \textbf{Avg-Initial} method is the most effective compared to \textbf{Re-Initial} and \textbf{Local-Initial}. Compared to \textbf{Re-Initial}, Avg-Initial ensures continuity in LoRA module training, preventing the risk of uploading suboptimal client updates to the server. Compared to \textbf{Local-Initial}, it effectively disseminates global information across LoRA module training, promoting better knowledge sharing. By averaging local LoRA modules, this method captures a representative update, smooths extreme deviations, and fosters a more stable and consistent training.

\section{LoRA-FAIR: Simple but Effective Solution}
\label{sec:method}
Building on the challenges outlined in previous sections, we propose a novel aggregation mechanism, LoRA-FAIR (shown in \cref{fig: lorafair}), designed to address both server-side aggregation bias and client-side initialization lag simultaneously. LoRA-FAIR employs a residual-based approach to refine the global model update. Rather than relying solely on the averaged LoRA matrices \( \mathbf{\bar{A}} \) and \( \mathbf{\bar{B}} \), LoRA-FAIR introduces a correction term for \( \mathbf{\bar{B}} \), denoted as the residual LoRA module \( \mathbf{\Delta B} \), to tackle both the server-side and client-side issues concurrently. Notably, LoRA-FAIR refines the global LoRA matrices at the server, without introducing additional communication or computational costs on the client side. In this section, we outline the key steps of LoRA-FAIR and demonstrate how it simultaneously addresses both Challenge 1 and Challenge 2.

To illustrate the process, consider a FL setup with \( K \) clients participating in fine-tuning at round \( t+1 \).

\textbf{Server Side.}
After fine-tuning in round \( t \), each client \( k \) sends its locally fine-tuned LoRA modules \( \mathbf{A}_k \) and \( \mathbf{B}_k \) back to the server. The server first aggregates these local modules to obtain the global modules \( \mathbf{\bar{A}} \) and \( \mathbf{\bar{B}} \) using \cref{global_modules}. Rather than directly distributing \( \mathbf{\bar{A}} \) and \( \mathbf{\bar{B}} \) to the clients, LoRA-FAIR refines the server-side aggregation by introducing a residual update \( \mathbf{\Delta B} \), optimizing the following:

\begin{equation} 
\label{eq:optimization_problem}
\arg\min\limits_{\mathbf{\Delta B}} \underbrace{\mathcal{S}\left(\mathbf{\Delta W}, (\mathbf{\bar{B}} + \mathbf{\Delta B}) \mathbf{\bar{A}}\right)}_{\text{correction}} + \underbrace{\lambda ||\mathbf{\Delta B}||}_{\text{regularization}},
\end{equation}
where \( \mathbf{\Delta W} \) represents the ideal global update from \cref{eq:ideal_update}, and \( \mathcal{S}(\cdot) \) is a similarity metric (i.e. cosine similarity~\cite{chowdhury2010introduction}) that measures the discrepancy between \( (\mathbf{\bar{B}} + \mathbf{\Delta B}) \mathbf{\bar{A}} \) and \( \mathbf{\Delta W} \). The regularization weight \( \lambda \) balances the correction term and the regularization term. We denote the corrected averaged LoRA \(\mathbf{B}\) with the residual as \(\mathbf{\bar{B}'} = \mathbf{\bar{B}} + \mathbf{\Delta B}\). The application of the residual update to LoRA \(\mathbf{\bar{B}}\) is validated through experiments and analysis in \cref{sec:ablation}. The optimization problem in \cref{eq:optimization_problem} can be approximately solved using SGD, with its computational cost detailed in the Appendix.

Upon determining \( \mathbf{\Delta B} \), the server distributes \(\mathbf{\bar{B}'} = \mathbf{\bar{B}} + \mathbf{\Delta B}\) and \( \mathbf{\bar{A}} \) to the clients for the next training round. This approach introduces no additional communication costs. Unlike existing methods that require large-matrix SVD computations \cite{bai2024federated} or transmission of all client-stacked LoRA modules \cite{wang2024flora}, LoRA-FAIR achieves computational and communication efficiency.

\textbf{Client Side.}  
Once client \( k \) receives $\mathbf{\bar{B}'}$ and \( \mathbf{\bar{A}} \), it begins local fine-tuning for round \( t+1 \) using its local dataset. The client initializes its LoRA module as \( \mathbf{B}_k = \mathbf{\bar{B}'} \) and \( \mathbf{A}_k = \mathbf{\bar{A}} \), while keeping the pre-trained model fixed.


\begin{table*}[h!]
\centering
\resizebox{0.85\textwidth}{!}{%
\begin{tabular}{c c c c c c c c c | c}

\toprule
\multirow{12}{*}{\textbf{DomainNet}} & \multirow{2}{*}{} & \multirow{2}{*}{} & \multirow{2}{*}{\textbf{Clipart}} & \multirow{2}{*}{\textbf{Infograph}} & \multirow{2}{*}{\textbf{Painting}} & \multirow{2}{*}{\textbf{Quickdraw}} & \multirow{2}{*}{\textbf{Real}}   & \multirow{2}{*}{\textbf{Sketch}} & \multirow{2}{*}{\textbf{Average}} \\
& & & & & & & & &\\
\cmidrule(r){2-10}
    & \multirow{6}{*}{\textbf{ViT}} & Centralized & 85.20 $\pm$ 0.018 & 57.15 $\pm$ 0.037 & 81.48 $\pm$ 0.014 & 73.09 $\pm$ 0.005 & 90.90 $\pm$ 0.003 & 78.81 $\pm$ 0.036 & 77.77 \\
    \cmidrule(r){3-10}
    & & FFA-LoRA & 81.75 $\pm$ 0.038 & 51.96 $\pm$ 0.058 & 77.51 $\pm$ 0.029 & 61.83 $\pm$ 0.095 & 88.68 $\pm$ 0.011 & 75.20 $\pm$ 0.050 & 72.82 \\
    &   & FedIT & 84.37 $\pm$ 0.069 & 54.17 $\pm$ 0.127 & 79.67 $\pm$ 0.047 & 69.00 $\pm$ 0.085 & 89.20 $\pm$ 0.012 & 78.08 $\pm$ 0.035 & 75.75 \\
    &   & FLoRA & 83.70 $\pm$ 0.041 & 53.51 $\pm$ 0.075 & 79.43 $\pm$ 0.046 & 70.09 $\pm$ 0.046 & 89.25 $\pm$ 0.011 & 77.20 $\pm$ 0.060 & 75.53 \\
    &   & FlexLoRA & 85.15 $\pm$ 0.034 & 53.93 $\pm$ 0.132 & 79.82 $\pm$ 0.034 & 70.01 $\pm$ 0.058 & 89.42 $\pm$ 0.010 & 77.85 $\pm$ 0.048 & 76.02 \\
    &   & \cellcolor{gray!40} \textbf{LoRA-FAIR}& \cellcolor{gray!40}\textbf{86.25 $\pm$ 0.032} & \cellcolor{gray!40}\textbf{56.26 $\pm$ 0.062} & \cellcolor{gray!40}\textbf{80.09 $\pm$ 0.072} & \cellcolor{gray!40}\textbf{71.25 $\pm$ 0.039} & \cellcolor{gray!40}\textbf{89.52 $\pm$ 0.014} & \cellcolor{gray!40}\textbf{79.06 $\pm$ 0.061} & \cellcolor{gray!40}\textbf{77.07} \\ 
\cline{2-10}
    & \multirow{6}{*}{\textbf{MLP-Mixer}} & Centralized & 74.61 $\pm$ 0.020 & 43.27 $\pm$ 0.019 & 71.54 $\pm$ 0.048 & 58.13 $\pm$ 0.039 & 85.90 $\pm$ 0.005 & 66.40 $\pm$ 0.048 & 66.64 \\
    \cmidrule(r){3-10}
    & & FFA-LoRA & 69.74 $\pm$ 0.021 & 37.15 $\pm$ 0.045 & 66.43 $\pm$ 0.018 & 38.66 $\pm$ 0.081 & 80.94 $\pm$ 0.006 & 57.49 $\pm$ 0.047 & 58.40 \\
    &     & FedIT & 74.69 $\pm$ 0.074 & 41.89 $\pm$ 0.089 & \textbf{70.57 $\pm$ 0.029} & 51.53 $\pm$ 0.030 & 83.25 $\pm$ 0.007 & 64.31 $\pm$ 0.130 & 64.37 \\
    &     & FLoRA & 74.39 $\pm$ 0.024 & 41.33 $\pm$ 0.072 & 69.91 $\pm$ 0.021 & 53.83 $\pm$ 0.039 & 82.75 $\pm$ 0.008 & 64.08 $\pm$ 0.017 & 64.38 \\
    &     & FlexLoRA & 75.11 $\pm$ 0.039 & 41.62 $\pm$ 0.146 & 70.49 $\pm$ 0.033 & 53.29 $\pm$ 0.051 & 83.41 $\pm$ 0.006 & 64.79 $\pm$ 0.028 & 64.79 \\
    &     &  \cellcolor{gray!40}\textbf{LoRA-FAIR} & \cellcolor{gray!40}\textbf{75.92 $\pm$ 0.039} & \cellcolor{gray!40}\textbf{43.21 $\pm$ 0.104} & \cellcolor{gray!40}70.42 $\pm$ 0.089 & \cellcolor{gray!40}\textbf{55.62 $\pm$ 0.041} & \cellcolor{gray!40}\textbf{83.43 $\pm$ 0.011} & \cellcolor{gray!40}\textbf{66.62 $\pm$ 0.039} & \cellcolor{gray!40}\textbf{65.87} \\
\midrule
\multirow{12}{*}{\textbf{NICO++}} & \multirow{2}{*}{} & \multirow{2}{*}{} & \multirow{2}{*}{\textbf{Autumn}} & \multirow{2}{*}{\textbf{Dim}} & \multirow{2}{*}{\textbf{Grass}} & \multirow{2}{*}{\textbf{Outdoor}} & \multirow{2}{*}{\textbf{Rock}}   & \multirow{2}{*}{\textbf{Water}} & \multirow{2}{*}{\textbf{Average}} \\
& & & & & & & & &\\
\cmidrule(r){2-10}
    & \multirow{6}{*}{\textbf{ViT}} & Centralized & 92.74 $\pm$ 0.063 & 89.63 $\pm$ 0.059 & 93.93 $\pm$ 0.024 & 91.07 $\pm$ 0.074 & 90.96 $\pm$ 0.036 & 90.71 $\pm$ 0.054 & 91.51 \\
    \cmidrule(r){3-10}
    &   & FFA-LoRA & 91.26 $\pm$ 0.019 & 88.19 $\pm$ 0.053 & 93.29 $\pm$ 0.012 & 89.84 $\pm$ 0.024 & 90.51 $\pm$ 0.019 & 88.60 $\pm$ 0.048 & 90.28 \\
    &   & FedIT & 91.64 $\pm$ 0.024 & 88.87 $\pm$ 0.047 & 93.09 $\pm$ 0.015 & 90.05 $\pm$ 0.028 & 90.87 $\pm$ 0.075 & 88.96 $\pm$ 0.029 & 90.58 \\
    &   & FLoRA & 91.48 $\pm$ 0.043 & 89.47 $\pm$ 0.063 & 93.33 $\pm$ 0.037 & 90.38 $\pm$ 0.040 & 90.83 $\pm$ 0.041 & 90.05 $\pm$ 0.057 & 90.93 \\
    &   & FlexLoRA & 91.26 $\pm$ 0.065 & 88.91 $\pm$ 0.042 & 93.16 $\pm$ 0.013 & 90.41 $\pm$ 0.026 & 90.78 $\pm$ 0.029 & 89.09 $\pm$ 0.043 & 90.60 \\
    &   &\cellcolor{gray!40} \textbf{LoRA-FAIR} & \cellcolor{gray!40}\textbf{92.47 $\pm$ 0.032} & \cellcolor{gray!40}\textbf{89.35 $\pm$ 0.054} & \cellcolor{gray!40}\textbf{93.73 $\pm$ 0.016} & \cellcolor{gray!40}\textbf{90.56 $\pm$ 0.025} & \cellcolor{gray!40}\textbf{91.01 $\pm$ 0.060} & \cellcolor{gray!40}\textbf{90.34 $\pm$ 0.035} & \cellcolor{gray!40}\textbf{91.24} \\ 
\cline{2-10}
    & \multirow{6}{*}{\textbf{MLP-Mixer}} & Centralized & 86.59 $\pm$ 0.042 & 82.15 $\pm$ 0.072 & 87.75 $\pm$ 0.012 & 83.67 $\pm$ 0.025  & 84.25 $\pm$ 0.037 & 82.60 $\pm$ 0.036 & 84.50 \\
    \cmidrule(r){3-10}
    &     & FFA-LoRA & 83.34 $\pm$ 0.015 & 76.82 $\pm$ 0.030 & 84.70 $\pm$ 0.010 & 80.14 $\pm$ 0.016 & 79.30 $\pm$ 0.008 & 75.97 $\pm$ 0.023 & 80.05 \\
    &     & FedIT & 85.21 $\pm$ 0.021 & 79.62 $\pm$ 0.066 & 86.01 $\pm$ 0.010 & 82.44 $\pm$ 0.031 & 83.10 $\pm$ 0.075 & 78.65 $\pm$ 0.026 & 82.51 \\
    &     & FLoRA & 85.10 $\pm$ 0.029 & 79.70 $\pm$ 0.068 & 86.03 $\pm$ 0.031 & 82.12 $\pm$ 0.055 & 82.24 $\pm$ 0.024 & 75.52 $\pm$ 0.023 & 82.29 \\
    &     & FlexLoRA & 86.31 $\pm$ 0.082 & 79.82 $\pm$ 0.051 & 86.60 $\pm$ 0.012 & 82.77 $\pm$ 0.023 & 83.05 $\pm$ 0.012 & 79.73 $\pm$ 0.045 & 83.08 \\
    &     & \cellcolor{gray!40} \textbf{LoRA-FAIR} & \cellcolor{gray!40}\textbf{86.09 $\pm$ 0.037} & \cellcolor{gray!40}\textbf{81.06 $\pm$ 0.048} & \cellcolor{gray!40}\textbf{86.79 $\pm$ 0.022} & \cellcolor{gray!40}\textbf{82.71 $\pm$ 0.018} & \cellcolor{gray!40}\textbf{84.09 $\pm$ 0.033} & \cellcolor{gray!40}\textbf{80.60 $\pm$ 0.033} & \cellcolor{gray!40}\textbf{83.56} \\
\bottomrule
\end{tabular}%
}
\caption{\textbf{Performance comparison} with baselines across different domains on DomainNet and NICO++ datasets using ViT and MLP-Mixer models in a \textbf{feature non-IID setting}. \textbf{Average} means the average accuracy across all domains. See details in \cref{sec:exp}.}
\label{tab:performance_feature}
\vspace{-10pt}
\end{table*}

\subsection{LoRA-FAIR for Challenge 1} LoRA-FAIR tackles the server-side aggregation bias by introducing the residual correction term \( \mathbf{\Delta B} \), which refines the aggregated LoRA matrix \( \mathbf{\bar{B}} \) on the server. In contrast to straightforward averaging, which leads to \( \mathbf{\bar{B}} \mathbf{\bar{A}} \) diverging from the ideal global update \( \mathbf{\Delta W} = \sum_{k=1}^K p_k \mathbf{B}_k \mathbf{A}_k \), LoRA-FAIR computes a residual update that minimizes the difference between the aggregated update and the ideal. By optimizing \( \mathbf{\Delta B} \), LoRA-FAIR approximates the target global model update more accurately, reducing the bias introduced by direct averaging. This correction ensures that the server-generated update better captures the interactions between local LoRA matrices, aligning \( (\mathbf{\bar{B}} + \mathbf{\Delta B}) \mathbf{\bar{A}} \) with the true aggregated update.

\subsection{LoRA-FAIR for Challenge 2} LoRA-FAIR also addresses the client-side initialization lag by adopting the principle of \textbf{Avg-Initial}. Specifically, $\mathbf{W} \gets \mathbf{W}$,  $\mathbf{A} \gets \bar{\mathbf{A}}$, $\mathbf{B} \gets \bar{\mathbf{B}} + \Delta \mathbf{B}$. The regularization term in LoRA-FAIR’s objective function prevents $\mathbf{\bar{B}'}$  from deviating excessively from \( \mathbf{\bar{B}} \), thus preserving the global average information obtained from the previous round. This approach maintains continuity between rounds, allowing clients to build upon a stable and consistent initialization that incorporates both local updates and global insights. By incorporating this regularization, LoRA-FAIR fosters a smoother transition and more effective local fine-tuning.

\section{Experiments}

\textbf{Foundation Models.} This paper primarily utilizes two foundation models commonly applied in computer vision (CV) tasks. \textbf{ViT} \cite{dosovitskiy2020image}: We use a pre-trained Vision Transformer (ViT) model with 12 transformer layers as a foundation model, pre-trained on ImageNet-21k \cite{deng2009imagenet} (specifically, “vit base patch16 224”). \textbf{MLP-Mixer} \cite{tolstikhin2021mlp}: In addition to ViT, we also use the MLP-Mixer model with 12 layers, pre-trained on ImageNet-21k, specifically “mixer b16 224”.
We follow the step in \cite{su2023fedra} for fine-tuning and the rank of LoRA is set as 16 for experiments.

\noindent \textbf{Datasets.} We conduct experiments on two real-world image datasets to simulate real client data distributions. \textbf{DomainNet} \cite{peng2019moment}: DomainNet is a large multi-domain dataset containing around 600k images across 345 categories, distributed over six domains: clipart, infograph, painting, quickdraw, real, and sketch. Following the setup in \cite{su2023fedra}, we use the first 100 categories. \textbf{NICO++} \cite{he2021towards}: NICO++ is an enhanced version of NICO dataset, containing approximately 90k images across 60 categories, representing six styles: autumn, dim, grass, outdoor, rock, and water.

To emulate real client data distribution, we focus on the \textbf{feature non-IID} setting, where each client has data from different domains. In this setting, we simulate six clients, each associated with one of the six distinct domains. Additionally, we conduct experiments under the \textbf{feature and label non-IID} setting, where we consider 30 clients in total, with each domain distributed among five clients. Label non-IID conditions among the five clients from each domain are generated using a Dirichlet distribution \cite{li2022federated} with a concentration parameter of 0.5.

\noindent \textbf{Training Details.}
The reported results are averaged over three independent runs. We use a mini-batch size of 128 and set the number of local iterations to 2 in feature non-IID setting and 5 in feature and label nonIID setting. We set the global rounds as 50 and 30 for DomainNet and NICO++ datasets respectively. The learning rate for local training is set to 0.01, with SGD as the optimizer. In the feature non-IID experiments, all 6 clients participate in the training. For the feature and label non-IID experiments, we consider that 18 clients participate in each communication round to simulate a partial participation setting.

\noindent \textbf{Baselines.} To evaluate the performance of our proposed method, LoRA-FAIR, we compare it with several state-of-the-art methods in federated fine-tuning with LoRA. \textbf{1. FedIT}: FedIT \cite{zhang2024towards} is the earliest approach to integrate LoRA with FedAvg. \textbf{2. FFA-LoRA}: FFA-LoRA \cite{sun2024improving} addresses server-side aggregation bias by fixing matrix \(\mathbf{A}\) and fine-tuning only matrix \(\mathbf{B}\). \textbf{3. FLoRA}: FLoRA \cite{wang2024flora} stacks local LoRA modules and transmits the stacked modules to all participating clients to mitigate server-side aggregation bias. \textbf{4. FlexLoRA}: FlexLoRA \cite{bai2024federated} reformulates each client’s local LoRA modules into a local update, sums these updates to generate a global update, and then applies SVD to update the local LoRA modules. \textbf{5. Centralized}: We also include an ideal centralized LoRA fine-tuning setting, where all data are held by a single entity, serving as a potential upper bound for comparison.

\subsection{Experiments Results}
\begin{table*}[h!]
\centering
\resizebox{0.85\textwidth}{!}{%
\begin{tabular}{c c c c c c c c c | c}

\toprule
\multirow{12}{*}{\textbf{DomainNet}} & \multirow{2}{*}{} & \multirow{2}{*}{} & \multirow{2}{*}{\textbf{Clipart}} & \multirow{2}{*}{\textbf{Infograph}} & \multirow{2}{*}{\textbf{Painting}} & \multirow{2}{*}{\textbf{Quickdraw}} & \multirow{2}{*}{\textbf{Real}}   & \multirow{2}{*}{\textbf{Sketch}} & \multirow{2}{*}{\textbf{Average}} \\
& & & & & & & & &\\
\cmidrule(r){2-10}
    & \multirow{6}{*}{\textbf{ViT}} & Centralized & 85.20 $\pm$ 0.018 & 57.15 $\pm$ 0.037 & 81.48 $\pm$ 0.014 & 73.09 $\pm$ 0.005 & 90.90 $\pm$ 0.003 & 78.81 $\pm$ 0.036 & 77.77 \\
    \cmidrule(r){3-10}
    & & FFA-LoRA & 81.75 $\pm$ 0.018 & 51.96 $\pm$ 0.022 & 77.51 $\pm$ 0.051 & 61.83 $\pm$ 0.025 & 88.68 $\pm$ 0.106 & 75.20 $\pm$ 0.091 & 72.82 \\
    &   & FedIT & 84.08 $\pm$ 0.029 & 52.94 $\pm$ 0.024 & 79.62 $\pm$ 0.067 & 61.03 $\pm$ 0.019 & 88.94 $\pm$ 0.074 & 76.70 $\pm$ 0.125 & 73.89 \\
    &   & FLoRA & 83.97 $\pm$ 0.042 & 53.57 $\pm$ 0.043 & 80.01 $\pm$ 0.063 & 62.77 $\pm$ 0.058 & 88.95 $\pm$ 0.060 & 76.30 $\pm$ 0.082 & 74.26 \\
    &   & FlexLoRA & 84.29 $\pm$ 0.018 & 53.60 $\pm$ 0.036 & 79.54 $\pm$ 0.084 & 62.05 $\pm$ 0.079 & 89.23 $\pm$ 0.055 & 76.76 $\pm$ 0.085 & 74.25 \\
    &   & \cellcolor{gray!40}\textbf{LoRA-FAIR}& \cellcolor{gray!40}\textbf{84.99 $\pm$ 0.024} & \cellcolor{gray!40}\textbf{55.15 $\pm$ 0.058} & \cellcolor{gray!40}\textbf{80.51 $\pm$ 0.038} & \cellcolor{gray!40}\textbf{62.77 $\pm$ 0.059} & \cellcolor{gray!40}\textbf{89.48 $\pm$ 0.026} & \cellcolor{gray!40}\textbf{77.03 $\pm$ 0.053} & \cellcolor{gray!40}\textbf{74.99} \\ 
\cline{2-10}
    & \multirow{6}{*}{\textbf{MLP-Mixer}} & Centralized & 74.61 $\pm$ 0.020 & 43.27 $\pm$ 0.019 & 71.54 $\pm$ 0.048 & 58.13 $\pm$ 0.039 & 85.90 $\pm$ 0.005 & 66.40 $\pm$ 0.048 & 66.64 \\
    \cmidrule(r){3-10}
    & & FFA-LoRA & 62.91 $\pm$ 0.025 & 33.65 $\pm$ 0.062 & 64.47 $\pm$ 0.022 & 25.76 $\pm$ 0.024 & 79.85 $\pm$ 0.040 & 50.63 $\pm$ 0.017 & 52.88 \\
    &     & FedIT & 71.53 $\pm$ 0.072 & 39.00 $\pm$ 0.089 & 68.76 $\pm$ 0.084 & 42.44 $\pm$ 0.060 & 82.34 $\pm$ 0.021 & 60.58 $\pm$ 0.071 & 60.77 \\
    &     & FLoRA & 70.06 $\pm$ 0.052 & 37.26 $\pm$ 0.095 & 67.48 $\pm$ 0.095 & 41.56 $\pm$ 0.090 & 81.37 $\pm$ 0.016 & 60.01 $\pm$ 0.106 & 59.62 \\
    &     & FlexLoRA & 71.58 $\pm$ 0.058 & 39.50 $\pm$ 0.033 & 68.89 $\pm$ 0.024 & 43.85 $\pm$ 0.091 & 82.39 $\pm$ 0.012 & 60.99 $\pm$ 0.096 & 61.20 \\
    &     & \cellcolor{gray!40}\textbf{LoRA-FAIR} & \cellcolor{gray!40}\textbf{72.79 $\pm$ 0.013} & \cellcolor{gray!40}\textbf{40.91 $\pm$ 0.043} & \cellcolor{gray!40}\textbf{69.49 $\pm$ 0.064} & \cellcolor{gray!40}\textbf{45.99 $\pm$ 0.073} & \cellcolor{gray!40}\textbf{82.59 $\pm$ 0.054} & \cellcolor{gray!40}\textbf{61.91 $\pm$ 0.059} & \cellcolor{gray!40}\textbf{62.28} \\
\midrule
\multirow{12}{*}{\textbf{NICO++}} & \multirow{2}{*}{} & \multirow{2}{*}{} & \multirow{2}{*}{\textbf{Autumn}} & \multirow{2}{*}{\textbf{Dim}} & \multirow{2}{*}{\textbf{Grass}} & \multirow{2}{*}{\textbf{Outdoor}} & \multirow{2}{*}{\textbf{Rock}}   & \multirow{2}{*}{\textbf{Water}} & \multirow{2}{*}{\textbf{Average}} \\
& & & & & & & & &\\
\cmidrule(r){2-10}
    & \multirow{6}{*}{\textbf{ViT}} & Centralized & 92.74 $\pm$ 0.063 & 89.63 $\pm$ 0.059 & 93.93 $\pm$ 0.024 & 91.07 $\pm$ 0.074 & 90.96 $\pm$ 0.036 & 90.71 $\pm$ 0.054 & 91.51 \\
    \cmidrule(r){3-10}
    & & FFA-LoRA & 91.42 $\pm$ 0.013 & 86.99 $\pm$ 0.056 & 92.06 $\pm$ 0.045 & 88.83 $\pm$ 0.048 & 90.10 $\pm$ 0.065 & 87.29 $\pm$ 0.035 & 89.45 \\
    &   & FedIT & 91.31 $\pm$ 0.035 & 86.91 $\pm$ 0.029 & 92.33 $\pm$ 0.008 & 89.01 $\pm$ 0.093 & 89.97 $\pm$ 0.042 & 87.37 $\pm$ 0.068 & 89.48 \\
    &   & FLoRA & 91.28 $\pm$ 0.105 & 87.07 $\pm$ 0.062 & 92.27 $\pm$ 0.020 & 89.52 $\pm$ 0.064 & 90.04 $\pm$ 0.056 & 87.43 $\pm$ 0.019 & 89.60 \\
    &   & FlexLoRA & \textbf{91.81} $\pm$ 0.031 & 87.23 $\pm$ 0.046 & 92.45 $\pm$ 0.007 & 89.25 $\pm$ 0.021 & 89.79 $\pm$ 0.033 & 87.37 $\pm$ 0.071 & 89.65 \\
    &   &\cellcolor{gray!40} \textbf{LoRA-FAIR} & \cellcolor{gray!40}91.79 $\pm$ 0.040 & \cellcolor{gray!40}\textbf{87.59 $\pm$ 0.060} & \cellcolor{gray!40}\textbf{92.90 $\pm$ 0.016} & \cellcolor{gray!40}\textbf{89.98 $\pm$ 0.028} & \cellcolor{gray!40}\textbf{90.39 $\pm$ 0.025} & \cellcolor{gray!40}\textbf{87.60 $\pm$ 0.018} & \cellcolor{gray!40}\textbf{90.04} \\ 
\cline{2-10}
    & \multirow{6}{*}{\textbf{MLP-Mixer}} & Centralized & 86.59 $\pm$ 0.042 & 82.15 $\pm$ 0.072 & 87.75 $\pm$ 0.012 & 83.67 $\pm$ 0.025  & 84.25 $\pm$ 0.037 & 82.60 $\pm$ 0.036 & 84.50 \\
    \cmidrule(r){3-10}
    & & FFA-LoRA & 80.04 $\pm$ 0.075 & 72.98 $\pm$ 0.048 & 82.07 $\pm$ 0.044 & 77.68 $\pm$ 0.081 & 76.23 $\pm$ 0.051 & 71.65 $\pm$ 0.068 & 76.78 \\
    &     & FedIT & 81.47 $\pm$ 0.021 & 74.50 $\pm$ 0.092 & 83.64 $\pm$ 0.075 & 78.67 $\pm$ 0.084 & 78.72 $\pm$ 0.019 & 74.20 $\pm$ 0.081 & 78.53 \\
    &     & FLoRA & 80.92 $\pm$ 0.018 & 74.58 $\pm$ 0.055 & 83.15 $\pm$ 0.027 & 79.21 $\pm$ 0.053 & 78.36 $\pm$ 0.035 & 74.25 $\pm$ 0.122 & 78.41 \\
    &     & FlexLoRA & 82.02 $\pm$ 0.032 & 75.02 $\pm$ 0.015 & 83.33 $\pm$ 0.021 & 78.88 $\pm$ 0.012 & 78.94 $\pm$ 0.030 & 74.25 $\pm$ 0.098 & 78.73 \\
    &     & \cellcolor{gray!40} \textbf{LoRA-FAIR} & \cellcolor{gray!40}\textbf{82.46 $\pm$ 0.049} & \cellcolor{gray!40}\textbf{76.02 $\pm$ 0.027} & \cellcolor{gray!40}\textbf{83.79 $\pm$ 0.040} & \cellcolor{gray!40}\textbf{79.84 $\pm$ 0.041} & \cellcolor{gray!40}\textbf{80.16 $\pm$ 0.042} & \cellcolor{gray!40}\textbf{74.90 $\pm$ 0.091} & \cellcolor{gray!40}\textbf{79.53} \\
\bottomrule
\end{tabular}%
}
\caption{\textbf{Performance comparison} with baselines across different domains on DomainNet and NICO++ datasets using ViT and MLP-Mixer models in a \textbf{feature and label non-IID setting}. \textbf{Average} means the average accuracy across all domains. See details in \cref{sec:exp}.}
\label{tab: both_noniid}
\vspace{-10pt}
\end{table*}

\label{sec:exp}

\noindent\textbf{Performance Comparisons.} 
We first compare the performance of the global model across different domains under the \textbf{feature non-IID} setting. In \cref{tab:performance_feature}, we present the results of our proposed method, LoRA-FAIR, alongside baseline methods on the DomainNet and NICO++ datasets across each domain using ViT as the foundation model. FFA-LoRA, despite reducing computation costs and addressing server-side aggregation bias by fixing the LoRA module \(\mathbf{A}\), achieves the lowest performance due to limited parameter flexibility, as only \(\mathbf{B}\) is fine-tuned, constraining optimization capacity. The state-of-the-art baseline method, FLoRA, which addresses server-side aggregation bias by stacking and transmitting local LoRA modules to each client, also underperforms compared to LoRA-FAIR. Although FLoRA effectively transmits the exact server aggregation update to clients, it even shows comparable performance to FedIT, a basic combination of FedAVG and LoRA, on the DomainNet dataset with ViT. These observations underscore the importance of client initialization, as discussed in Challenge 2, where the starting point of client models significantly affects federated fine-tuning results. FlexLoRA, which uses SVD to decompose summed local updates, performs better than other baselines but still falls short of LoRA-FAIR. Our proposed method which considers both server-side aggregation bias and client initialization lag, achieves superior performance in individual domain assessments and overall average accuracy. Additional experiments on both datasets using the MLP-Mixer model show similar performance trends, further supporting our findings.

We then conduct experiments under the \textbf{feature and label non-IID} setting to further validate our proposed method. In this setup, we consider a total of 30 clients, with each group of 5 clients sharing the same data domain but having non-IID label distributions (using a Dirichlet distribution with a concentration parameter of 0.5). To simulate partial participation, we increase the number of local iterations to 5 and allow 18 clients to participate in each communication round. Results in \cref{tab: both_noniid} indicate that, even in this more challenging setting, our proposed method, LoRA-FAIR, continues to outperform the baseline methods.

\begin{figure*}[ht]
    \centering
    \begin{minipage}[b]{0.32\textwidth}
        \includegraphics[width=\linewidth, height=4cm]{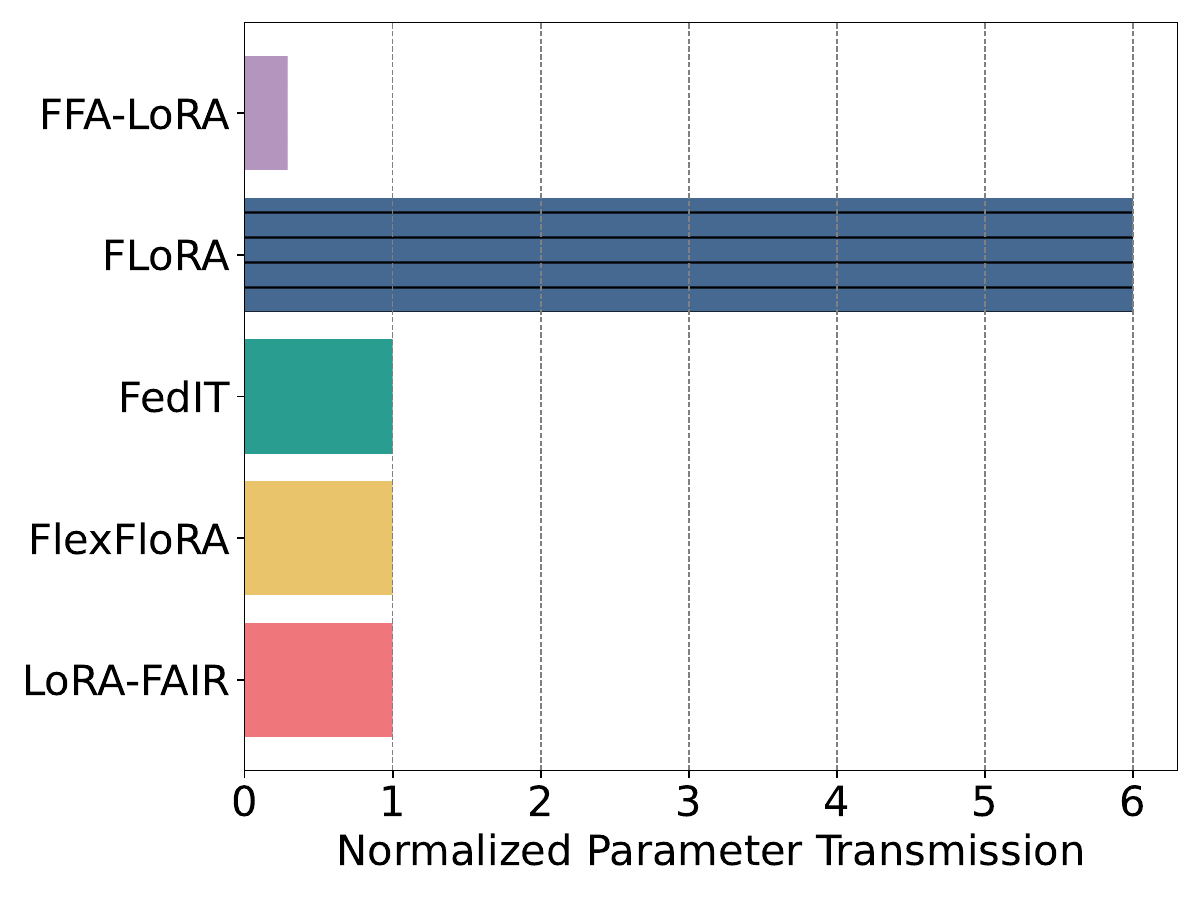} 
        \vspace{-20pt}
        \caption{\textbf{Communication cost comparison.} LoRA-FAIR matches the communication cost of FedIT and FlexLoRA and avoids FLoRA's high overhead. Details in \cref{sec:exp}.}
        \label{fig:communication}
    \end{minipage}
    \hspace{0.01\textwidth}
    \begin{minipage}[b]{0.32\textwidth}
        \includegraphics[width=\linewidth, height=4cm]{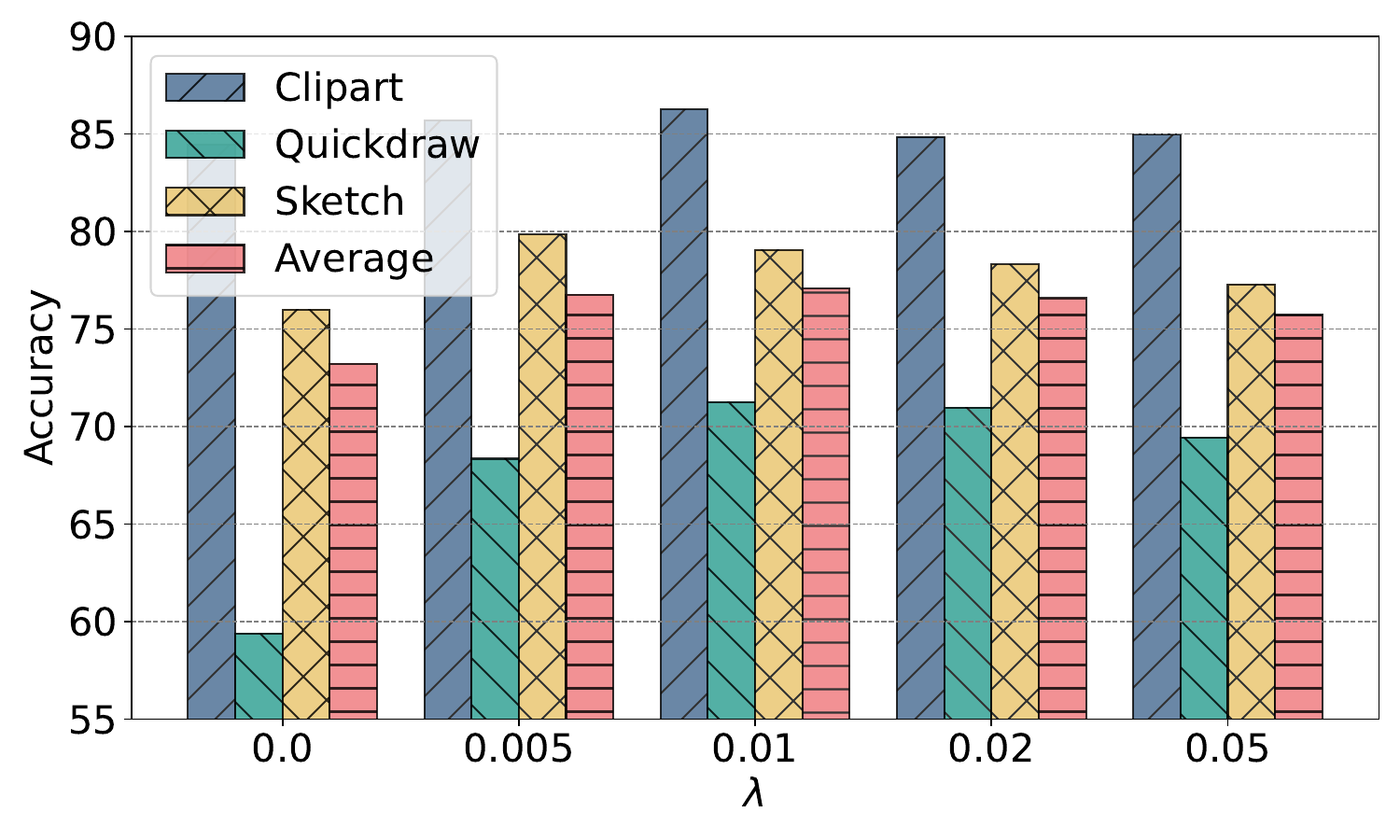} 
        \vspace{-20pt}
        \caption{\textbf{Impact of Regularization Weight $\lambda$.} With $\lambda = 0$, LoRA-FAIR results in the lowest performance, underscoring the importance of this term. Details in \cref{sec:ablation}.}
        \label{fig:lambda}
    \end{minipage}
    \hspace{0.01\textwidth}
    \begin{minipage}[b]{0.32\textwidth}
        \includegraphics[width=\linewidth, height=4cm]{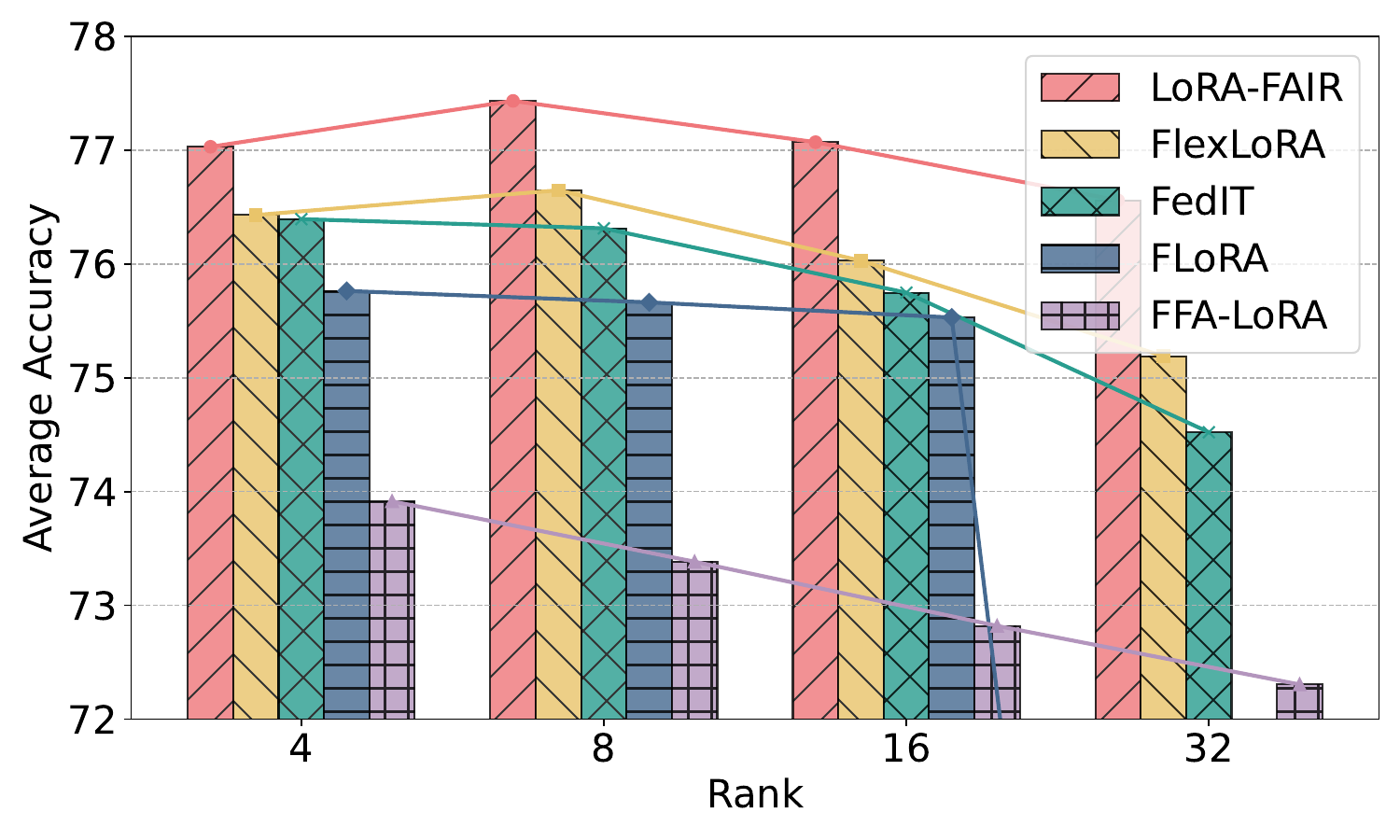} 
         \vspace{-20pt}
        \caption{\textbf{Impact of LoRA Rank.} LoRA-FAIR outperforms baselines across ranks \{4, 8, 16, 32\}, with higher ranks not always improving performance, consistent with \cite{cho2023heterogeneous}.}
        \label{fig:rank1}
    \end{minipage}
    \vspace{-20pt}
\end{figure*}

\noindent\textbf{Communication Overhead.} Here, we analyze the communication efficiency of our proposed method. As shown in \cref{fig:communication}, LoRA-FAIR only requires the server to distribute $\mathbf{\bar{B}'}$ and \( \mathbf{\bar{A}} \) to the clients each round, incurring no additional communication cost compared to FedIT and FlexLoRA. In contrast, FLoRA, which stacks all clients' local LoRA modules and distributes them to all clients, introduces significant communication overhead. FFA-LoRA has the lowest communication cost since it keeps the LoRA module \(\mathbf{A}\) fixed and only transmits \(\mathbf{B}\) each round. However, as shown in \cref{tab:performance_feature} and \cref{tab: both_noniid}, FFA-LoRA performs the worst across all settings. These results demonstrate that our proposed method achieves the best trade-off between communication cost and fine-tuned model performance.

\subsection{Ablation Studies}
\label{sec:ablation}
\textbf{Impact of Residual LoRA Module Position.}  
In our proposed method, we apply the residual update $\mathbf{\Delta B}$ to the LoRA module $\mathbf{B}$. To examine the impact of this choice, we conduct an ablation study by adding the residual update (denoted as $\mathbf{\Delta A}$) to the LoRA module $\mathbf{A}$ or applying the residual update to both LoRA modules, $\mathbf{A}$ and $\mathbf{B}$. This study is conducted on the DomainNet dataset using ViT as the foundation model. As shown in \cref{tab: residual_position}, adding the residual update to the LoRA module $\mathbf{B}$ achieves slightly better performance than the others. This finding aligns with the observation in~\cite{tian2025hydralora} that LoRA modules serve distinct functions, where $\mathbf{A}$ primarily captures general information and benefits from stability with averaged updates.

\begin{table}[h!]
\centering
\resizebox{\columnwidth}{!}{%
\begin{tabular}{c|cccccc|c}
\toprule
    Residual  & Clipart & Infograph & Painting & Quickdraw & Real  & Sketch & Average \\
\midrule
$\mathbf{\Delta A}$ & 84.93 & 54.55 & 80.08 & 71.13 & 89.48 & 78.39 & 76.42 \\
$\mathbf{\Delta A, \Delta B}$ & 84.69 & 54.64 & 78.56 & 68.73 & 88.28 & 78.39 & 75.55 \\
\rowcolor{gray!40} \textbf{$\mathbf{\Delta B}$} & \textbf{86.25} & \textbf{56.26} & \textbf{80.09} & \textbf{71.25} & \textbf{89.52} & \textbf{79.06} & \textbf{77.07} \\
\bottomrule
\end{tabular}%
}
\caption{\textbf{Performance comparison under different choices of residual LoRA modules position}. See details in \cref{sec:ablation}.}
\label{tab: residual_position}
\vspace{-10pt}
\end{table}


\noindent \textbf{Impact of Regularization Weight $\lambda$.} 
In our proposed method, we optimize the objective in \cref{eq:optimization_problem} to address both server aggregation bias and client initialization lag. Notably, we include a regularization term \( \lambda ||\mathbf{\Delta B}|| \) to balance the similarity measure with the correction term. Here, we conduct experiments to investigate the impact of the regularization weight $\lambda$ on model performance. As shown in \cref{fig:lambda}, varying $\lambda$ affects the performance of LoRA-FAIR, highlighting the importance of this parameter. Specifically, when $\lambda = 0$, LoRA-FAIR achieves its lowest performance. 


\begin{table}[h]
\centering
\resizebox{\columnwidth}{!}{%
\begin{tabular}{c|ccc}
\toprule
    Regularization Term \( \lambda ||\mathbf{\Delta B}|| \)  & $\mathcal{S}(\mathbf{\bar{B}},\mathbf{\bar{B}} + \mathbf{\Delta B})$ & $\mathcal{S}\left(\mathbf{\Delta W},(\mathbf{\bar{B}} + \mathbf{\Delta B})\mathbf{\bar{A}}\right)$ & Average Accuracy \\
\midrule
w/o ($\lambda = 0$) & 0.971488 & \textbf{0.999847} & 73.22 \\
\rowcolor{gray!40} \textbf{w/ ($\lambda = 0.01$)} & \textbf{0.999808} & 0.999701 & \textbf{77.07}\\
\bottomrule
\end{tabular}%
}
\caption{\textbf{Impact of the regularization term on the similarity and the average accuracy metrics.} See details in \cref{sec:ablation}.}
\label{tab: cos}
\vspace{-10pt}
\end{table}

This occurs because, as shown in \cref{tab: cos}, while setting $\lambda = 0$ helps address server aggregation bias by approximating $(\mathbf{\bar{B}} + \mathbf{\Delta B}) \mathbf{\bar{A}}$ to $\mathbf{\Delta W}$, it reduces the similarity between $(\mathbf{\bar{B}} + \mathbf{\Delta B})$ and $\mathbf{\bar{B}}$, failing to mitigate client initialization lag. This result highlights the significant role of client initialization in influencing model performance. Additionally, with small regularization values (e.g., $\lambda = 0.01, 0.02$), performance remains stable. Thus, we recommend setting the regularization weight to a small positive value. In our experimental setup, we set the regularization weight to 0.01.


\noindent \textbf{Impact of LoRA Rank.} 
In this subsection, we investigate the impact of different LoRA ranks by conducting experiments with ranks set to \{4, 8, 16, 32\}. Notably, FLoRA fails to converge when the rank is 32, highlighting the limitations of its approach, which involves direct updates to the pre-trained model. We observe that increasing the LoRA rank does not necessarily lead to better final performance, consistent with findings from previous studies \cite{cho2023heterogeneous}. Additionally, the results in \cref{fig:rank1} demonstrate that our proposed method consistently outperforms baselines across all rank settings, validating its effectiveness.

Additional experiment results on convergence performance, adaptation to clients with heterogeneous LoRA ranks, server-side computational overhead, and the limitation of FLoRA can be found in the Appendix.

\section{Related Work}

\noindent \textbf{Parameter-Efficient Fine-Tuning.}  
The increasing size of foundation models makes full fine-tuning computationally and storage-intensive. To address these challenges, Parameter-Efficient Fine-Tuning (PEFT) methods \cite{fu2023effectiveness, ding2023parameter, liu2022few, han2024parameter} have been proposed to reduce the number of trainable parameters. PEFT techniques introduce a limited set of additional trainable parameters to enhance model performance while keeping most pre-trained parameters frozen. Some approaches, such as \cite{houlsby2019parameter}, add trainable parameters called adapters to each layer of the pre-trained network, updating only the adapters during fine-tuning. Other approaches, like \cite{cai2020tinytl}, focus on fine-tuning only the bias terms of the pre-trained model. Techniques such as prefix-tuning \cite{li2021prefix} and prompt-tuning \cite{lester2021power} add trainable dimensions to the input or hidden layers of the network. Among PEFT methods, a key approach is LoRA \cite{hu2021lora}, which uses low-rank matrices to approximate the pre-trained weight matrix, updating only the low-rank matrices. In this paper, we focus on LoRA due to its demonstrated efficiency, achieving comparable performance to full-parameter fine-tuning.

\noindent\textbf{Federated Learning.}  
FedAvg \cite{mcmahan2017communication}, the foundational work in FL, demonstrates the advantages of this approach in terms of privacy and communication efficiency by aggregating local model parameters to train a shared global model. Numerous FL studies \cite{mcmahan2017communication, wang2024aggregation, yang2024fedfed, wang2024taming, wang2024federated, bian2024prioritizing, peng2024fedmm, liu2024fedbcgd, liuimproving} have addressed various challenges within FL settings. For example, several works explore the impact of different initialization strategies on model performance. \cite{tan2022federated} shows that initializing with pre-trained weights can enhance the stability of FedAvg's global aggregation, while \cite{tian2022fedbert} confirms the effectiveness of using a pre-trained model as an initial starting point. However, these methods primarily focus on smaller models and do not extend to foundation models or incorporate parameter-efficient fine-tuning; instead, they adhere to conventional FL training practices.


\noindent\textbf{Federated Fine-Tuning.}  
Several studies \citep{cho2023heterogeneous, kuang2024federatedscope, wu2024fedbiot, sun2024improving, wang2024flora, bai2024federated} have explored federated fine-tuning approaches. For example, \citet{kuang2024federatedscope} proposes federated fine-tuning with all parameters updated, while \citet{sun2022conquering} introduces federated fine-tuning with PEFT using prefix-tuning. A closely related area to our work involves federated fine-tuning using LoRA. \citet{zhang2024towards, zhang2023fedpetuning} apply LoRA in a federated context; however, these methods overlook potential server aggregation bias. Several subsequent works have been proposed: FFA-LoRA \citep{sun2024improving} freezes the non-zero initialized low-rank matrices and updates only the zero-initialized matrices, FlexLoRA \citep{bai2024federated} uses SVD to redistribute weights, and FLoRA \citep{wang2024flora} stacks local LoRA modules and transmits them to each client. However, these methods do not address client initialization lag. A more detailed discussion of related federated LoRA works can be found in Appendix \cref{sec:prior works}.

\section{Conclusion}
\label{sec:conclusion}
In this work, we proposed LoRA-FAIR to address the key challenges of server-side aggregation bias and client-side initialization lag in federated fine-tuning with LoRA. LoRA-FAIR approximates an ideal solution by maintaining shared average information while ensuring dynamic server-side adjustments. Our experiments on large-scale datasets demonstrated its superior performance over state-of-the-art methods. Future work will explore extending LoRA-FAIR beyond computer vision datasets and adapting it for scenarios where clients use different LoRA ranks to enhance its applicability in diverse federated learning environments.


\newpage
\section{Acknowledgments}
The work of J. Bian, L. Wang and J. Xu is partially supported by NSF under grants 2505381 and 2515982. The work of L. Zhang is partially supported by NSF under grant 2348279.

{
    \small
    \bibliographystyle{ieeenat_fullname}
    \bibliography{main}
}

\clearpage
\setcounter{page}{1}
\maketitlesupplementary

\section{Additional Experiments Results}
\label{sec :appendix_exp}
In this section, we provide additional experimental details and results to further validate our proposed method, LoRA-FAIR.

\subsection{Convergence Performance}

\begin{figure}[ht]
    \begin{minipage}[b]{0.23\textwidth}
	\centering
   	\includegraphics[width=\linewidth]{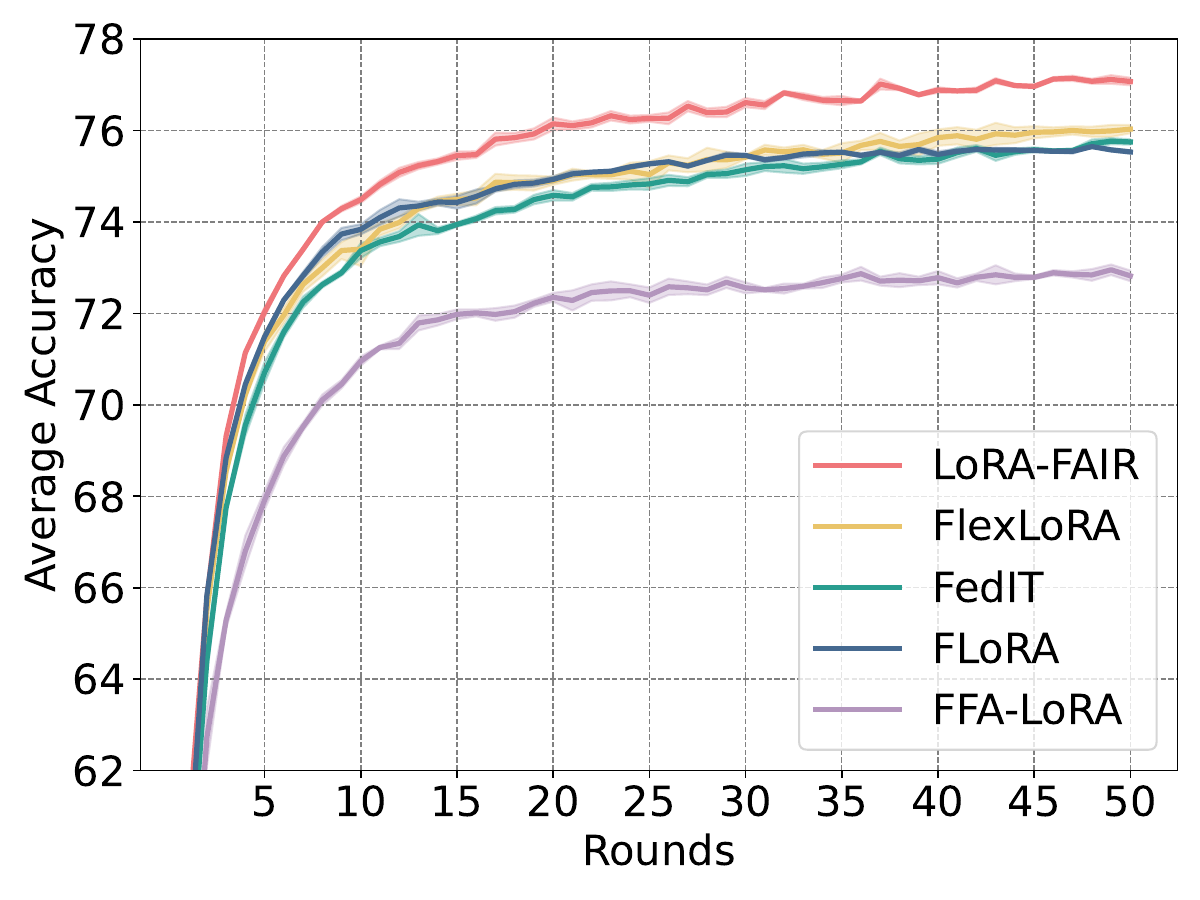}
	\end{minipage}
    \hspace{0.005\textwidth}
    \begin{minipage}[b]{0.23\textwidth}
	\centering\includegraphics[width=\linewidth]{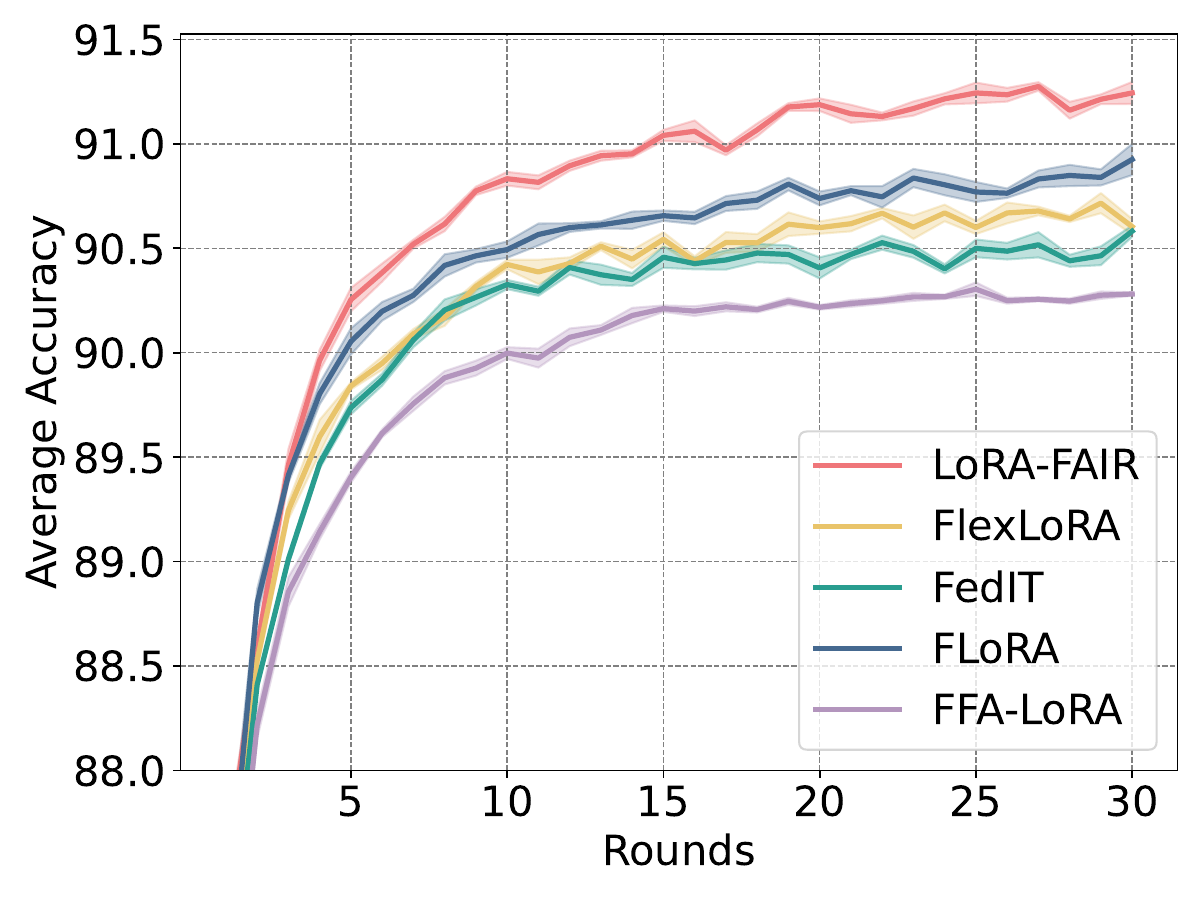}
	\end{minipage}
    \caption{\textbf{Comparison of average accuracy} across training rounds on DomainNet (\textbf{left}) and NICO++ (\textbf{right}) datasets using the ViT model. The shaded area indicates the variance across multiple runs. For more details, refer to \cref{sec:exp}.}
    \label{fig:performance}
    \vspace{-10pt}
\end{figure}

\begin{figure}[ht]
    \begin{minipage}[b]{0.23\textwidth}
	\centering 	\includegraphics[width=\linewidth]{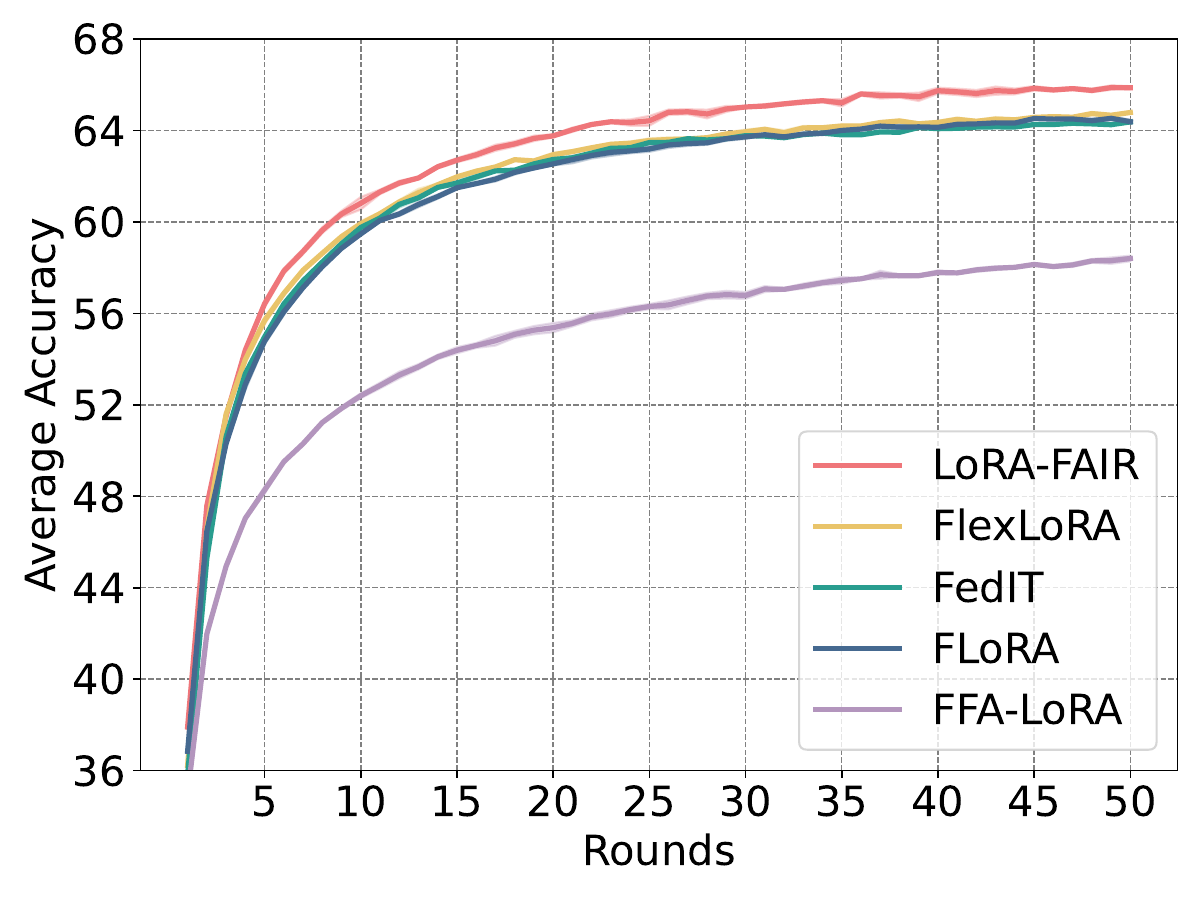}
	\end{minipage}
    \hspace{0.005\textwidth}
    \begin{minipage}[b]{0.23\textwidth}
	\centering\includegraphics[width=\linewidth]{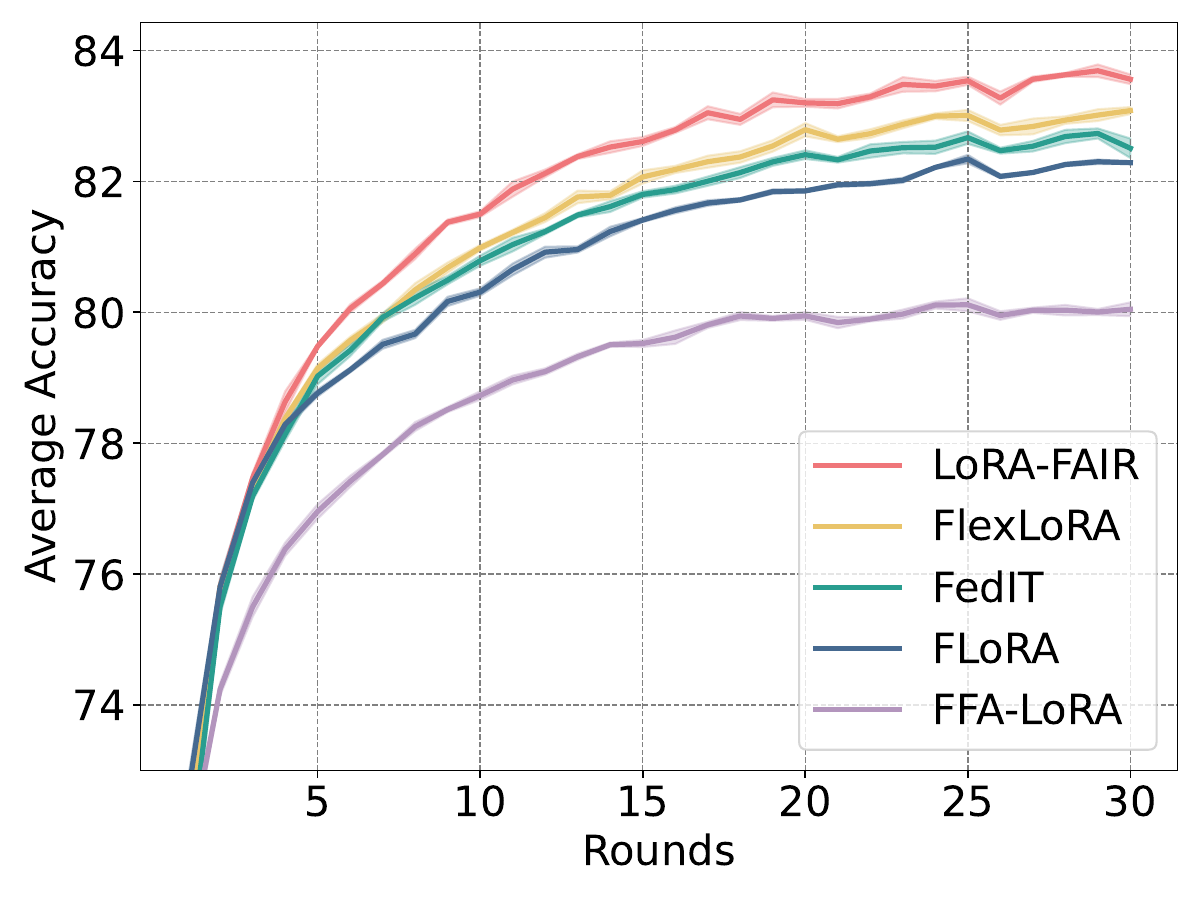}
	\end{minipage}
    \caption{\textbf{Comparison of average accuracy} across training rounds on DomainNet (\textbf{left}) and NICO++ (\textbf{right}) datasets using the Mixer model. The shaded area indicates the variance across multiple runs. For more details, refer to \cref{sec:exp}.}
    \label{fig:performance_mixer}
    \vspace{-10pt}
\end{figure}

We present the convergence performance of our proposed method compared to baseline methods using the ViT or MLP-Mixer model under feature non-IID setting. As shown in \cref{fig:performance} and \cref{fig:performance_mixer}, our proposed method consistently outperforms all baseline methods. These results are consistent with those in the main paper (\cref{tab:performance_feature}), further validating the robustness of our approach.

\begin{table*}[h!]
\centering
\resizebox{0.85\textwidth}{!}{%
\begin{tabular}{c c c c c c c  | c}

\toprule
 \multirow{2}{*}{\textbf{Method}} & \multirow{2}{*}{\textbf{Clipart}} & \multirow{2}{*}{\textbf{Infograph}} & \multirow{2}{*}{\textbf{Painting}} & \multirow{2}{*}{\textbf{Quickdraw}} & \multirow{2}{*}{\textbf{Real}}   & \multirow{2}{*}{\textbf{Sketch}} & \multirow{2}{*}{\textbf{Average}} \\
& & & & & & &\\
\cmidrule(r){0-7}
    HETLoRA & 73.96 & 42.57 & 74.49 & 27.52 & 87.05 & 59.74 & 60.89 \\
    FlexLoRA & 83.11 & 52.43 & 78.63 & 62.30 & 88.23 & 77.32 & 73.50 \\
    \cellcolor{gray!40}\textbf{LoRA-FAIR + HETLoRA}& \cellcolor{gray!40}\textbf{83.40} & \cellcolor{gray!40}\textbf{52.25} & \cellcolor{gray!40}\textbf{79.28} &\cellcolor{gray!40}\textbf{63.24} & \cellcolor{gray!40}\textbf{89.40} & \cellcolor{gray!40}\textbf{77.74} & \cellcolor{gray!40}\textbf{74.22} \\ 
\bottomrule
\end{tabular}%
}
\caption{\textbf{Performance comparison} with baselines across different domains on DomainNet using ViT model with client having heterogeneous LoRA rank. \textbf{Average} means the average accuracy across all domains. See details in \cref{sec: heter_rank}.}
\label{tab: heter_rank}
\vspace{-10pt}
\end{table*}

\begin{table}[h!]
\centering
\resizebox{\columnwidth}{!}{%
\begin{tabular}{c|c|cccccc|c|c}
\toprule
    Method & Local Epoch & Clipart & Infograph & Painting & Quickdraw & Real  & Sketch & Average & $\Delta$ \\
\midrule
FLoRA & 2 & 85.15 & 53.51 & 79.43 & 70.09 & 89.25 & 77.20 & 75.53 & -\\
Proposed & 2 & 86.25 & 56.26 & 80.09 & 71.25 & 89.52 & 79.06 & 77.07 & +1.54\\
\midrule
FLoRA & 10 & 83.81 & 52.91 & 78.36 & 61.25 & 88.68 & 76.57 & 73.60 & -\\
Proposed & 10 & 85.20 & 53.39 & 79.03 & 61.51 & 89.24 & 77.47 & 74.31 & +0.71 \\
\bottomrule
\end{tabular}%
}
\caption{\textbf{Limitation of FLoRA's Reinitialization}. FLoRA's reinitialization strategy fails to learn an optimal client update under smaller local training, leading to suboptimal model performance. See \cref{sec:local_epochs} for details.}
\label{tab: local_epoch}
\vspace{-10pt}
\end{table}

\subsection{Adaptation for Clients with Heterogeneous LoRA Ranks}
\label{sec: heter_rank}
Our proposed method primarily focuses on settings where clients have the same LoRA rank, addressing challenges such as server aggregation bias and client initialization lag when combining LoRA with federated learning. However, our approach can be extended to scenarios where clients have heterogeneous LoRA ranks. 

The state-of-the-art method for handling heterogeneous ranks in FL is HETLoRA \cite{cho2023heterogeneous}, which employs zero-padding and truncation for distribution. It is important to note that HETLoRA is specifically designed for heterogeneous settings and operates orthogonally to our proposed method. By integrating zero-padding and truncation for distribution into LoRA-FAIR, our method can effectively operate in heterogeneous rank settings. We evaluate this adaptation using the DomainNet dataset with ViT as the foundation model. The client data distribution and training settings are consistent with those used in the feature non-IID experiments in the main paper. The clients LoRA ranks are set as $\{2, 4, 4, 6, 6, 8\}$. The results, presented in \cref{tab: heter_rank}, demonstrate that our proposed method, combined with zero-padding and truncation, achieves the best performance compared to existing methods, validating its effectiveness in heterogeneous rank scenarios. We note that due to the heterogeneous LoRA ranks, FedIT and FFA-LoRA are not suitable for this setting and are therefore excluded from the experiment. While FLoRA can operate under heterogeneous settings, it is not included in the results as it fails to converge in our experiments. This failure underscores its limitation of directly adding updates to the pre-trained model rather than updating the LoRA modules.

\subsection{Server-Side Computational Overhead Analysis}
\label{sec:computation}

\begin{figure}
    \centering
    \includegraphics[width=0.8\linewidth]{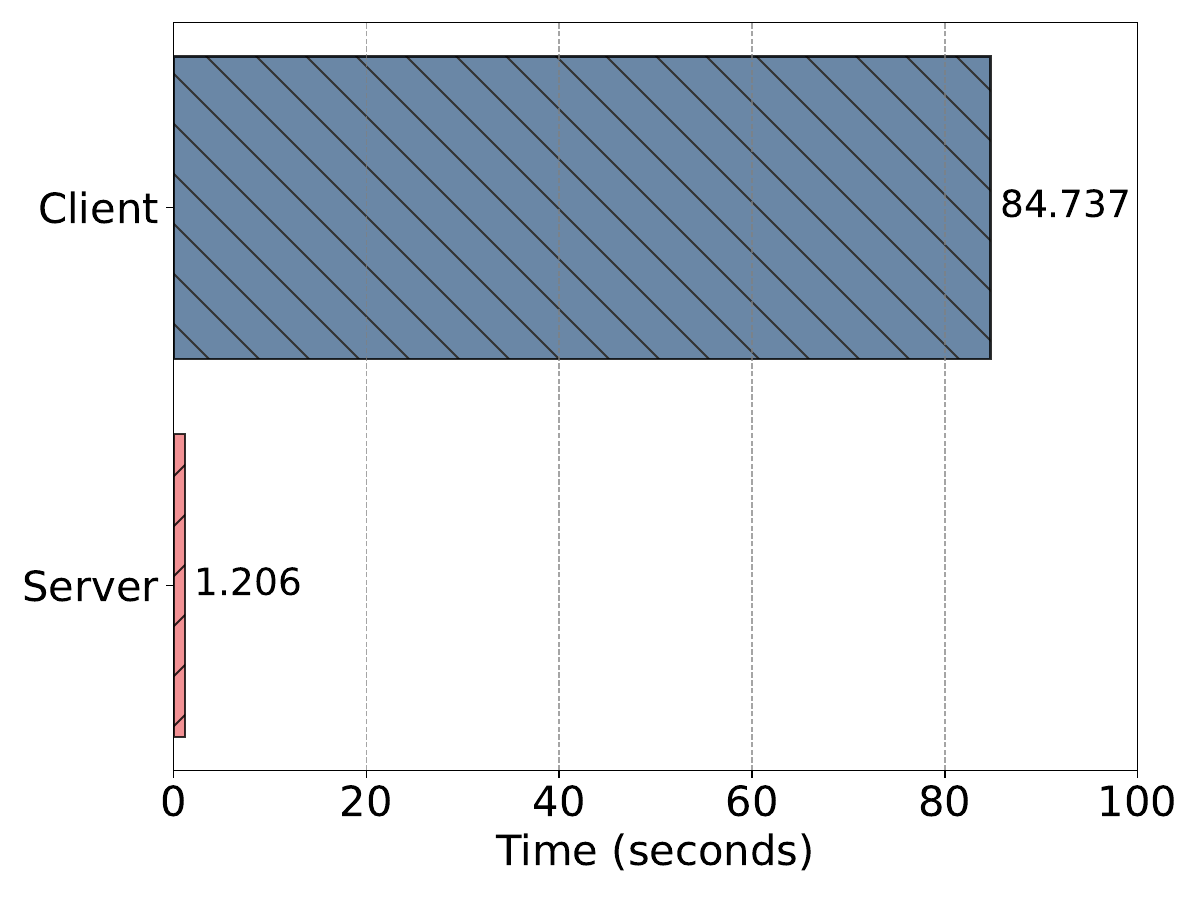}
    \caption{\textbf{Comparison of computational time} between the client and the server. See details in \cref{sec:computation}}
    \label{fig:computation}
\end{figure}

Our proposed method addresses both server aggregation bias and client initialization lag by solving \cref{eq:optimization_problem}, introducing only a small computational overhead on the server side. In our main experiment, we solve \cref{eq:optimization_problem} using SGD with a learning rate of 0.01 and 1000 iterations. However, this additional cost is minimal and can be considered negligible given the substantial computational resources typically available on servers. Moreover, a comparison of training times, as shown in \cref{fig:computation}, demonstrates that the time required to solve \cref{eq:optimization_problem} on the server is minimal compared to the client-side local training time.

\subsection{Limitation of FLoRA's Reinitialization}
\label{sec:local_epochs}

We evaluate performance under different local epochs. In our main experiments with the feature non-IID setting, we set the number of local epochs to 2 and the number of global rounds to 50. Here, we test a configuration where the local epochs are set to 10, and accordingly, the global rounds are reduced to 10 to maintain a fixed total number of updates. The results indicate that a shorter local epoch with more frequent updates leads to better performance for both the proposed method and FLoRA. This finding is consistent with \cite{zhao2018federated}, which suggests that the number of local epochs should not be too high in a non-IID FL setting. Furthermore, with shorter local epochs, the performance gap between our proposed method and FLoRA increases, further validating that FLoRA's reinitialization strategy fails to learn an optimal client update under limited local training, ultimately degrading the final model performance.

\section{Prior Works}
\label{sec:prior works}
In this section, we review existing methods and their limitations.

\noindent\textbf{FedIT \cite{zhang2024towards}}:  
FedIT is the earliest approach to integrate LoRA with FedAvg. In FedIT, each client starts with a fixed pre-trained foundation model and trains local LoRA modules, represented as low-rank matrices \( \mathbf{A}_k \) and \( \mathbf{B}_k \), on its private dataset. The server aggregates these local matrices into global LoRA modules through a weighted average based on data size. While computationally efficient, this method introduces server-side aggregation bias.

\noindent\textbf{FFA-LoRA \cite{sun2024improving}}:  
FFA-LoRA freezes the non-zero initialized low-rank matrix \(\mathbf{A}\) and updates only the zero-initialized matrix \(\mathbf{B}\). By freezing \(\mathbf{A}\), the actual global update becomes equal to the ideal global update (i.e., \(\mathbf{\Delta W} = \mathbf{\Delta W}'\)), addressing server-side aggregation bias. However, freezing \(\mathbf{A}\) significantly reduces the number of trainable parameters, limiting the model's capacity. Our experiments confirm that although FFA-LoRA resolves aggregation bias, its limited parameter flexibility results in worse performance compared to other baselines.

\noindent\textbf{FLoRA \cite{wang2024flora}}:  
FLoRA stacks local LoRA modules from all clients and transmits the stacked modules back to each client to reconstruct global updates, which are then added directly to each client's pre-trained model while reinitializing local LoRA modules for the next training round. Although FLoRA effectively addresses server-side aggregation bias, it incurs high communication costs proportional to the number of clients and raises privacy concerns, as it distributes all clients' LoRA modules rather than only the averaged ones. Additionally, FLoRA's reinitialization strategy introduces Client-Side Initialization Lag. Frequent reinitialization results in small gradient updates, leading to inefficient training and suboptimal performance.

\noindent\textbf{FlexLoRA \cite{bai2024federated}}:  
FlexLoRA reformulates each client’s local LoRA modules into a local update, sums these updates to generate a global update, and applies SVD to produce global LoRA modules. These modules are then distributed to clients as initialization for the next round. While this approach formulates an ideal global update, it still suffers from server-side aggregation bias due to the SVD step. For example, consider two clients, each with rank-8 LoRA modules (\( \text{Rank}(\Delta W_1) = 8 \) and \( \text{Rank}(\Delta W_2) = 8 \)), resulting in a global update with \(\text{Rank}(\Delta W) \leq 16\). Using SVD to produce global modules with a rank of 8 may lead to information loss, preventing the transmission of an ideal global update to clients.

\noindent\textbf{Comparison to Existing Efficient Weight Aggregation FL Methods.} Our work identifies a gap in existing federated learning methods concerning fine-tuning with LoRA. While prior approaches—such as layer-wise model aggregation~\cite{ma2022layer}, elastic aggregation~\cite{chen2023elastic}, and related layer-wise techniques~\cite{rehman2023dawa, lee2023layer}—have demonstrated effectiveness in general federated optimization, they are not well-suited for LoRA-based fine-tuning. Specifically, these methods do not address how to decompose aggregated model updates into the necessary LoRA modules for client model initialization, nor do they provide strategies to avoid the direct transmission of these updates. To overcome these limitations, our method tailors the aggregation process specifically for federated fine-tuning with LoRA, bridging the gap left by existing techniques.

\section{Theorem}

\begin{theorem} For analytical tractability, we consider the case where the similarity metric S is based on the Frobenius norm. The residual correction term $\Delta\mathbf{B}$ obtained by minimizing Equation (8) guarantees that $(\bar{\mathbf{B}}+\Delta\mathbf{B})\bar{\mathbf{A}}$ approaches the ideal global update $\Delta\mathbf{W}$ with the following approximation guarantee:

\begin{align}
&\|(\bar{\mathbf{B}}+\Delta\mathbf{B}^*)\bar{\mathbf{A}} - \Delta\mathbf{W}\|_F^2 \nonumber\\
\leq & \|\Delta\mathbf{W} - \bar{\mathbf{B}}\bar{\mathbf{A}}\|_F^2 \cdot \left(1 - \frac{\sigma_{\min}^2(\bar{\mathbf{A}})}{\sigma_{\min}^2(\bar{\mathbf{A}}) + \lambda}\right)^2,
\end{align}
here $\sigma_{\min}(\bar{\mathbf{A}})$ is the smallest non-zero singular value of $\bar{\mathbf{A}}$.
\end{theorem}

\begin{proof}
Let's denote $\mathbf{E} = \Delta\mathbf{W} - \bar{\mathbf{B}}\bar{\mathbf{A}}$ as the initial aggregation error. Our objective function becomes:

\begin{align}
J(\Delta\mathbf{B}) = \|\Delta\mathbf{B}\bar{\mathbf{A}} - \mathbf{E}\|_F^2 + \lambda\|\Delta\mathbf{B}\|_F^2
\end{align}

To find the critical points of $J(\Delta\mathbf{B})$, we take the derivative with respect to $\Delta\mathbf{B}$ and set it equal to zero:

\begin{align}
\nabla_{\Delta\mathbf{B}}J(\Delta\mathbf{B}) = 2(\Delta\mathbf{B}\bar{\mathbf{A}} - \mathbf{E})\bar{\mathbf{A}}^T + 2\lambda\Delta\mathbf{B} = 0
\end{align}

Solving for the optimal $\Delta\mathbf{B}^*$:
\begin{align}
\Delta\mathbf{B}^* = \mathbf{E}\bar{\mathbf{A}}^T(\bar{\mathbf{A}}\bar{\mathbf{A}}^T + \lambda\mathbf{I})^{-1},
\end{align}
where $\mathbf{I}$ is the identity matrix of appropriate dimensions. Substituting back the definition of $\mathbf{E}$:
\begin{align}
\Delta\mathbf{B}^* = (\Delta\mathbf{W} - \bar{\mathbf{B}}\bar{\mathbf{A}})\bar{\mathbf{A}}^T(\bar{\mathbf{A}}\bar{\mathbf{A}}^T + \lambda\mathbf{I})^{-1}
\end{align}

The residual error after applying the correction is:
\begin{align}
\mathbf{E}_{residual} & = -\mathbf{E} + \Delta\mathbf{B}^*\bar{\mathbf{A}} \nonumber\\
& = -\mathbf{E} + \mathbf{E}\bar{\mathbf{A}}^T(\bar{\mathbf{A}}\bar{\mathbf{A}}^T + \lambda\mathbf{I})^{-1}\bar{\mathbf{A}} \nonumber\\
& =  \mathbf{E}(-\mathbf{I} + \bar{\mathbf{A}}^T(\bar{\mathbf{A}}\bar{\mathbf{A}}^T + \lambda\mathbf{I})^{-1}\bar{\mathbf{A}})
\end{align}

Let's define the matrix $\mathbf{M} = -\mathbf{I} + \bar{\mathbf{A}}^T(\bar{\mathbf{A}}\bar{\mathbf{A}}^T + \lambda\mathbf{I})^{-1}\bar{\mathbf{A}}$. For any matrix $\bar{\mathbf{A}}$, the eigenvalues of $\bar{\mathbf{A}}^T(\bar{\mathbf{A}}\bar{\mathbf{A}}^T + \lambda\mathbf{I})^{-1}\bar{\mathbf{A}}$ can be bounded using the properties of matrix norms and the Sherman-Morrison-Woodbury formula:
\begin{align}
\bar{\mathbf{A}}^T(\bar{\mathbf{A}}\bar{\mathbf{A}}^T + \lambda\mathbf{I})^{-1}\bar{\mathbf{A}} = \bar{\mathbf{A}}^T\bar{\mathbf{A}}(\bar{\mathbf{A}}^T\bar{\mathbf{A}} + \lambda\mathbf{I})^{-1}
\end{align}
The eigenvalues of this matrix are of the form $\frac{\mu_i}{\mu_i + \lambda}$, where $\mu_i$ are the eigenvalues of $\bar{\mathbf{A}}^T\bar{\mathbf{A}}$. Since the eigenvalues of $\bar{\mathbf{A}}^T\bar{\mathbf{A}}$ are the squares of the singular values of $\bar{\mathbf{A}}$, i.e., $\mu_i = \sigma_i^2$, the eigenvalues of $\bar{\mathbf{A}}^T(\bar{\mathbf{A}}\bar{\mathbf{A}}^T + \lambda\mathbf{I})^{-1}\bar{\mathbf{A}}$ are $\frac{\sigma_i^2}{\sigma_i^2 + \lambda}$. Therefore, the eigenvalues of $\mathbf{M} = -\mathbf{I} + \bar{\mathbf{A}}^T(\bar{\mathbf{A}}\bar{\mathbf{A}}^T + \lambda\mathbf{I})^{-1}\bar{\mathbf{A}}$ are $-1 + \frac{\sigma_i^2}{\sigma_i^2 + \lambda} = -\frac{\lambda}{\sigma_i^2 + \lambda}$.

The spectral norm of $\mathbf{M}$ is the maximum absolute eigenvalue:

\begin{align}
\|\mathbf{M}\|_2 = \max_i \left|\frac{-\lambda}{\sigma_i^2 + \lambda}\right| = \frac{\lambda}{\sigma_{\min}^2 + \lambda}
\end{align}

Using the property that for any matrices $\mathbf{P}$ and $\mathbf{Q}$, $\|\mathbf{P}\mathbf{Q}\|_F \leq \|\mathbf{P}\|_F\|\mathbf{Q}\|_2$, we have:

\begin{align}
\|\mathbf{E}_{residual}\|_F &= \|\mathbf{E}\mathbf{M}\|_F \\
&\leq \|\mathbf{E}\|_F\|\mathbf{M}\|_2 \\
&= \|\mathbf{E}\|_F \cdot \frac{\lambda}{\sigma_{\min}^2 + \lambda}
\end{align}

Since $\frac{\lambda}{\sigma_{\min}^2 + \lambda} = 1 - \frac{\sigma_{\min}^2}{\sigma_{\min}^2 + \lambda}$, we have:

\begin{align}
\|\mathbf{E}_{residual}\|_F \leq \|\mathbf{E}\|_F \cdot \left(1 - \frac{\sigma_{\min}^2}{\sigma_{\min}^2 + \lambda}\right)
\end{align}
Squaring both sides and substituting $\mathbf{E} = \Delta\mathbf{W} - \bar{\mathbf{B}}\bar{\mathbf{A}}$, we get our final bound.
 
\end{proof}

\begin{corollary}
    When $\lambda = 0$ and $\bar{\mathbf{A}}$ has full row rank, there exists an exact solution where:
$$(\bar{\mathbf{B}}+\Delta\mathbf{B}^*)\bar{\mathbf{A}} = \Delta\mathbf{W}$$
\end{corollary}

\noindent As the regularization parameter $\lambda$ increases, the solution balances two objectives:
\begin{enumerate}
\item Minimizing the approximation error to the ideal update $\Delta\mathbf{W}$
\item Preventing large deviations from the averaged LoRA module $\bar{\mathbf{B}}$
\end{enumerate}


\begin{theorem}[Convergence of Federated LoRA Fine-Tuning]
\label{thm:convergence_improved}
Under standard FL assumptions (L-smooth loss, bounded gradients $G$, $E$ local epochs), and assuming the global learning rate $\eta$, the convergence of federated LoRA fine-tuning after $T$ rounds is:
\begin{align}
&\frac{1}{T}\sum_{t=0}^{T-1}\mathbb{E}[\|\nabla \mathcal{L}(W^t)\|^2] \leq \frac{4[\mathcal{L}(W^0) - \mathcal{L}(W^*)]}{\eta T} \nonumber\\
&+ 4\eta^2E^2G^2\left(\frac{L^2}{2} + 1\right)+ 8\frac{1}{T}\sum_{t=0}^{T-1}\|\Delta W^t - \bar{B}^t\bar{A}^t\|_F^2 \cdot \gamma\nonumber
\end{align}
where $\gamma$ characterizes the aggregation method:
1. For LoRA-FAIR: $\gamma = \left(1 - \frac{\sigma^2_{\min}(\bar{A}^t)}{\sigma^2_{\min}(\bar{A}^t) + \lambda}\right)^2 < 1$.
2. For FedIT (standard aggregation): $\gamma = 1$
\end{theorem}

\begin{proof}
At round $t$, the global model update is:
\begin{equation}
W^{t+1} = W^t - \eta \cdot \Delta W'^t
\end{equation}
where $\Delta W'^t = (\bar{B}^t + \Delta B^{*t})\bar{A}^t$ for LoRA-FAIR and $\Delta W'^t = \bar{B}^t\bar{A}^t$ for FedIT. The ideal global update is:
\begin{equation}
\Delta W^t = \sum_{k=1}^K p_k B_k^t A_k^t
\end{equation}

Define the aggregation error:
\begin{equation}
E_{agg}^t = \Delta W^t - \Delta W'^t
\end{equation}

From Theorem A.1, for LoRA-FAIR:
\begin{equation}
\|E_{agg}^t\|_F^2 \leq \|\Delta W^t - \bar{B}^t\bar{A}^t\|_F^2 \cdot \gamma
\end{equation}

Using L-smoothness:
\begin{align}
\mathcal{L}(W^{t+1}) &\leq \mathcal{L}(W^t) + \langle\nabla\mathcal{L}(W^t), W^{t+1} - W^t\rangle \nonumber\\
&\quad + \frac{L}{2}\|W^{t+1} - W^t\|^2\\
&= \mathcal{L}(W^t) - \eta\langle\nabla\mathcal{L}(W^t), \Delta W'^t\rangle \nonumber\\
&\quad + \frac{L\eta^2}{2}\|\Delta W'^t\|^2
\end{align}

Decomposing $\Delta W'^t = \Delta W^t - E_{agg}^t$:
\begin{align}
\langle\nabla\mathcal{L}(W^t), \Delta W'^t\rangle &= \langle\nabla\mathcal{L}(W^t), \Delta W^t\rangle \nonumber\\
&\quad - \langle\nabla\mathcal{L}(W^t), E_{agg}^t\rangle
\end{align}

Under bounded gradients:
\begin{equation}
\langle\nabla\mathcal{L}(W^t), \Delta W^t\rangle \geq \frac{1}{2}\|\nabla\mathcal{L}(W^t)\|^2 - \frac{\eta^2E^2L^2G^2}{2}
\end{equation}

For the error term:
\begin{equation}
\langle\nabla\mathcal{L}(W^t), -E_{agg}^t\rangle \leq \frac{1}{4}\|\nabla\mathcal{L}(W^t)\|^2 + \|E_{agg}^t\|^2
\end{equation}

Combining bounds:
\begin{align}
\mathcal{L}(W^{t+1}) &\leq \mathcal{L}(W^t) - \frac{\eta}{4}\|\nabla\mathcal{L}(W^t)\|^2 + \frac{\eta^3E^2L^2G^2}{2} \nonumber\\
&\quad + \eta\|E_{agg}^t\|^2 + \frac{L\eta^2}{2}\|\Delta W'^t\|^2
\end{align}

Using $\|\Delta W'^t\|^2 \leq 2(\|\Delta W^t\|^2 + \|E_{agg}^t\|^2)$ and the bound on $E_{agg}^t$:
\begin{align}
\mathcal{L}(W^{t+1}) &\leq \mathcal{L}(W^t) - \frac{\eta}{4}\|\nabla\mathcal{L}(W^t)\|^2 + \frac{\eta^3E^2L^2G^2}{2} \nonumber\\
&\quad + \eta\|E_{agg}^t\|^2 + \frac{L\eta^2}{2}(2(\|\Delta W^t\|^2 + \|E_{agg}^t\|^2))\nonumber\\
 &\leq \mathcal{L}(W^t) - \frac{\eta}{4}\|\nabla\mathcal{L}(W^t)\|^2 + \frac{\eta^3E^2L^2G^2}{2} \nonumber\\
&\quad + \eta\|E_{agg}^t\|^2 + {\eta}(\eta^2E^2G^2 + \|E_{agg}^t\|^2)\nonumber\\
 &\leq \mathcal{L}(W^t) - \frac{\eta}{4}\|\nabla\mathcal{L}(W^t)\|^2 + \frac{\eta^3E^2L^2G^2}{2} \nonumber\\
&\quad + 2\eta\|E_{agg}^t\|^2 + \eta^3E^2G^2
\end{align}
where $\eta <\frac{1}{L}$. Then we have:
\begin{align}
\mathcal{L}(W^{t+1}) &\leq \mathcal{L}(W^t) - \frac{\eta}{4}\|\nabla\mathcal{L}(W^t)\|^2 + \frac{\eta^3E^2L^2G^2}{2} \nonumber\\
&\quad + 2\eta\|\Delta W^t - \bar{B}^t\bar{A}^t\|_F^2 \cdot \gamma + \eta^3E^2G^2
\end{align}
Rearranging and taking expectation:
\begin{align}
\frac{\eta}{4}\mathbb{E}[\|\nabla\mathcal{L}(W^t)\|^2] &\leq \mathbb{E}[\mathcal{L}(W^t)] - \mathbb{E}[\mathcal{L}(W^{t+1})] \nonumber\\
&\quad + \frac{\eta^3E^2L^2G^2}{2} + \eta^3E^2G^2 \nonumber\\
&\quad + 2\eta\|\Delta W^t - \bar{B}^t\bar{A}^t\|_F^2 \cdot \gamma
\end{align}
Then by summing over t:
\begin{align}
\frac{\eta}{4}\sum_{t=0}^{T-1}\mathbb{E}[\|\nabla\mathcal{L}(W^t)\|^2] &\leq \mathcal{L}(W^0) - \mathbb{E}[\mathcal{L}(W^T)] \nonumber\\
&\quad + T\eta^3E^2G^2(\frac{L^2}{2} + 1) \nonumber\\
&\quad + 2\eta\sum_{t=0}^{T-1}\|\Delta W^t - \bar{B}^t\bar{A}^t\|_F^2 \cdot \gamma
\end{align}
dividing both sides by $\frac{\eta}{4}$ and T, we have:
\begin{align}
\frac{1}{T}\sum_{t=0}^{T-1}\mathbb{E}[\|\nabla\mathcal{L}(W^t)\|^2] &\leq \frac{4}{T\eta}[\mathcal{L}(W^0) - \mathcal{L}(W^*)] \nonumber\\
&\quad + 4\eta^2E^2G^2(\frac{L^2}{2} + 1) \nonumber\\
&\quad + 8\frac{1}{T}\sum_{t=0}^{T-1}\|\Delta W^t - \bar{B}^t\bar{A}^t\|_F^2 \cdot \gamma
\end{align}
With $\eta = \frac{1}{\sqrt{T}}$, we finally get:
\begin{align}
\frac{1}{T}\sum_{t=0}^{T-1}\mathbb{E}[\|\nabla\mathcal{L}(W^t)\|^2] &\leq \frac{4[\mathcal{L}(W^0) - \mathcal{L}(W^*)]}{\sqrt{T}} \nonumber\\
&\quad + 4\eta^2E^2G^2(\frac{L^2}{2} + 1) \nonumber\\
&\quad + 8\frac{1}{T}\sum_{t=0}^{T-1}\|\Delta W^t - \bar{B}^t\bar{A}^t\|_F^2 \cdot \gamma
\end{align}
\end{proof}
As data becomes more non-IID, $\|\Delta\mathbf{W} - \bar{\mathbf{B}}\bar{\mathbf{A}}\|_F$ increases, leading to a larger error term in standard methods. LoRA-FAIR reduces this impact by the factor $\gamma < 1$, providing tighter convergence bounds especially in highly non-IID settings.

\end{document}